\documentclass[runningheads]{llncs}
\usepackage{graphicx}
\usepackage{comment}
\usepackage{amsmath,amssymb} 
\usepackage{color}
\usepackage{caption}
\usepackage{subcaption}
\usepackage{booktabs}
\usepackage{makecell}
\usepackage{enumitem}
\usepackage{multirow}
\usepackage{microtype}
\usepackage{hyperref}

\newcommand{\boldparagraph}[1]{\vspace{0.2em}\noindent{\bf #1} }

\DeclareMathOperator*{\argmin}{arg\,min}
\begin{document}
\pagestyle{headings}
\mainmatter
\def\ECCVSubNumber{4158}

\title{Online Invariance Selection\\for Local Feature Descriptors}

\titlerunning{Online Invariance Selection for Local Feature Descriptors}
\author{R\'emi Pautrat${}^1$ \and
Viktor Larsson${}^1$ \and
Martin R. Oswald${}^1$ \and
Marc Pollefeys${}^{1, 2}$}
\authorrunning{R. Pautrat et al.}
\institute{${}^1$ Department of Computer Science, ETH Zurich \\ ${}^2$ Microsoft Mixed Reality and AI Zurich lab}

\maketitle

\begin{abstract}

To be invariant, or not to be invariant: that is the question formulated in this work about local descriptors. A limitation of current feature descriptors is the trade-off between generalization and discriminative power: more invariance means less informative descriptors. We propose to overcome this limitation with a disentanglement of invariance in local descriptors and with an online selection of the most appropriate invariance given the context. Our framework\footnote{\url{https://github.com/rpautrat/LISRD}} consists in a joint learning of multiple local descriptors with different levels of invariance and of meta descriptors encoding the regional variations of an image. The similarity of these meta descriptors across images is used to select the right invariance when matching the local descriptors. Our approach, named Local Invariance Selection at Runtime for Descriptors (LISRD), enables descriptors to adapt to adverse changes in images, while remaining discriminative when invariance is not required. We demonstrate that our method can boost the performance of current descriptors and outperforms state-of-the-art descriptors in several matching tasks, when evaluated on challenging datasets with day-night illumination as well as viewpoint changes.

\keywords{Local descriptors, invariance, visual localization}
\end{abstract}

\section{Introduction}
%
Sparse features detection and description is at the root of many computer vision tasks: Structure-from-Motion (SfM), Simultaneous Localization and Mapping (SLAM), image retrieval, tracking, etc. They offer a compact representation in terms of memory storage and allow for efficient image matching, and are thus well suited for large-scale applications~\cite{heinly2015_reconstructing_the_world,schnberger2017semantic,Sattler_2017_CVPR}. These features should however be able to cope with real world conditions such as day-night changes~\cite{Zhou2016W}, seasonal variations~\cite{Sattler_2018_CVPR} and matching across large baselines~\cite{Tola10daisyan}.

To be able to do matching in extreme scenarios, the successive feature detectors and descriptors have become more and more invariant~\cite{Mikolajczyk05acomparison}. The Harris corner detector~\cite{Harris88acombined} was already invariant to rotations, but not to scale. The SIFT detector and descriptor~\cite{lowe2004} was one of the first to achieve invariance with respect to scale, rotation and uniform light changes. More recently, learned descriptors have been able to encode invariance without handcrafting it. On the one hand, patch-based descriptors can become invariant to transforms when estimating the shape of the patch~\cite{yi2016,ono2018lfnet,Mishkin_2018_ECCV,Ebel_2019_ICCV}. On the other hand, recent dense descriptors leverage the power of large convolutional neural networks (CNN) to become more general and invariant. Most of them are trained on images with many variations in the training set, either obtained through data augmentation~\cite{detone2018}, with large databases of challenging images~\cite{Dusmanu2019CVPR,yang2020ur2kid} or with style transfer~\cite{revaud2019}. They can also directly encode the invariance in the network itself~\cite{liu2019}. The general trend in descriptor learning is thus to capture as much invariance as possible.

\begin{figure}[t]
    \centering
    \tiny
    \setlength{\tabcolsep}{0.3mm}
    \newcommand{\sz}{0.32}
    \begin{tabular}{cccc}
         & SIFT & Upright SIFT & Our method\\
        \rotatebox{90}{\hspace{1pt}With rotation} &
        \includegraphics[width=\sz\textwidth]{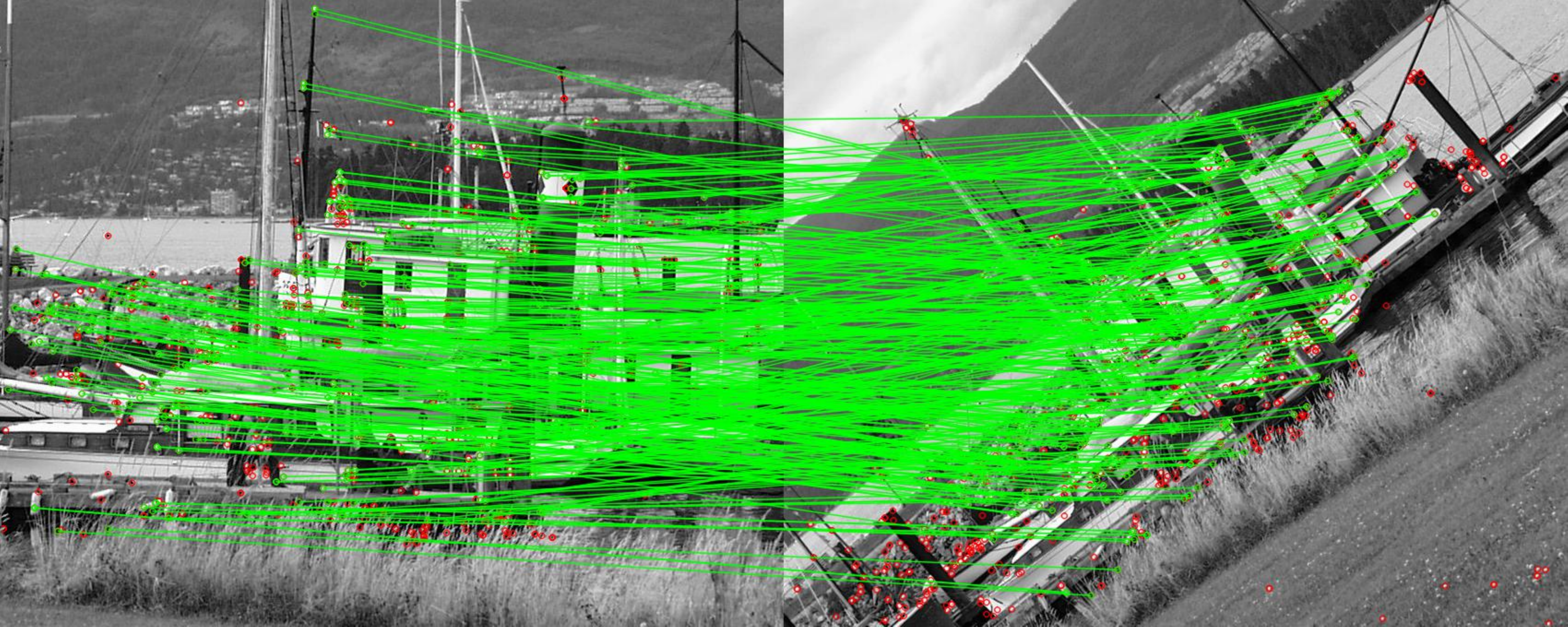} &
        \includegraphics[width=\sz\textwidth]{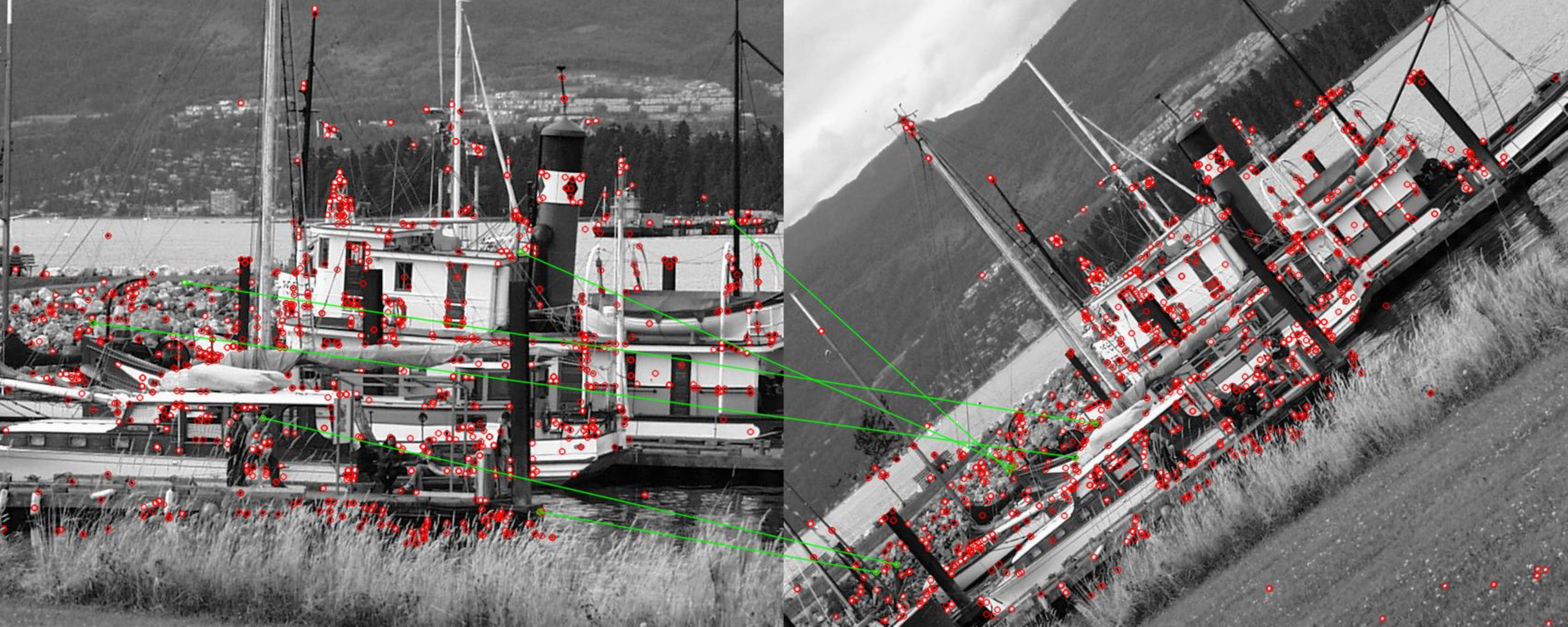} &
        \includegraphics[width=\sz\textwidth]{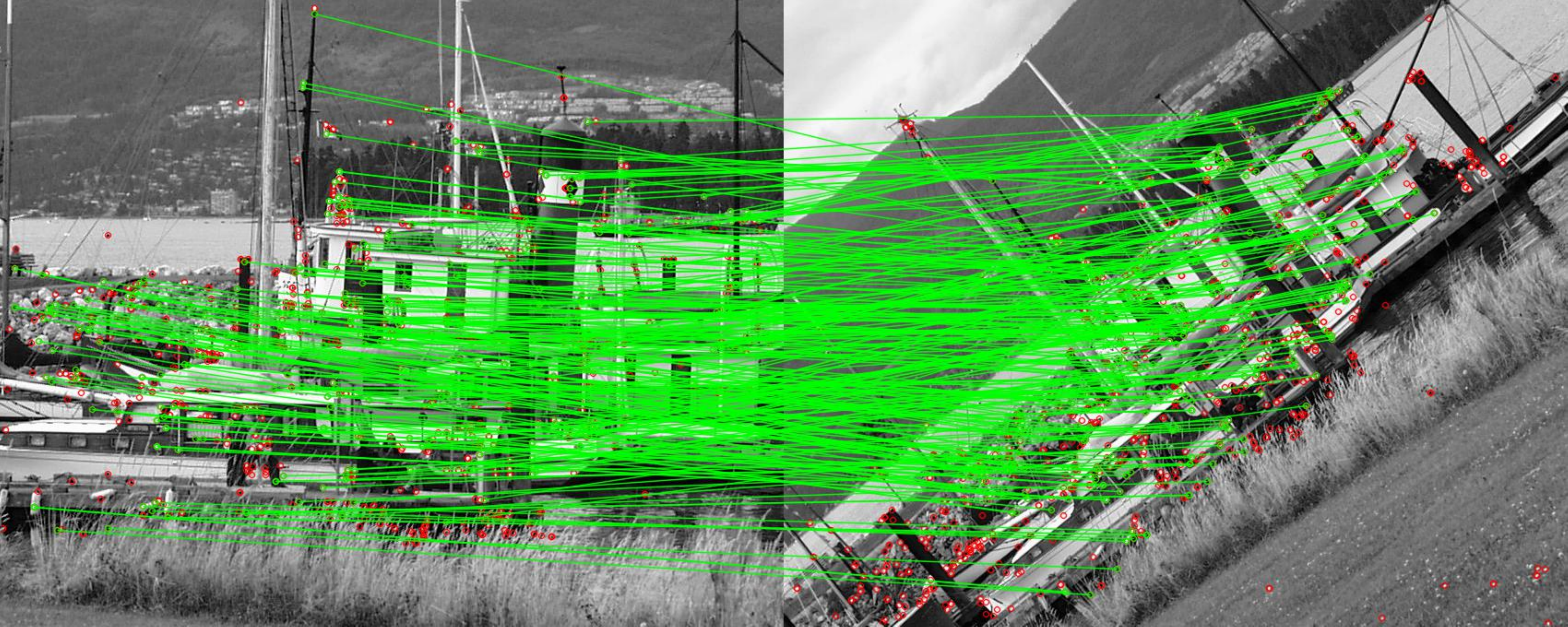}\\[-0.4mm]
        \rotatebox{90}{\hspace{4pt}No rotation} &
        \includegraphics[width=\sz\textwidth]{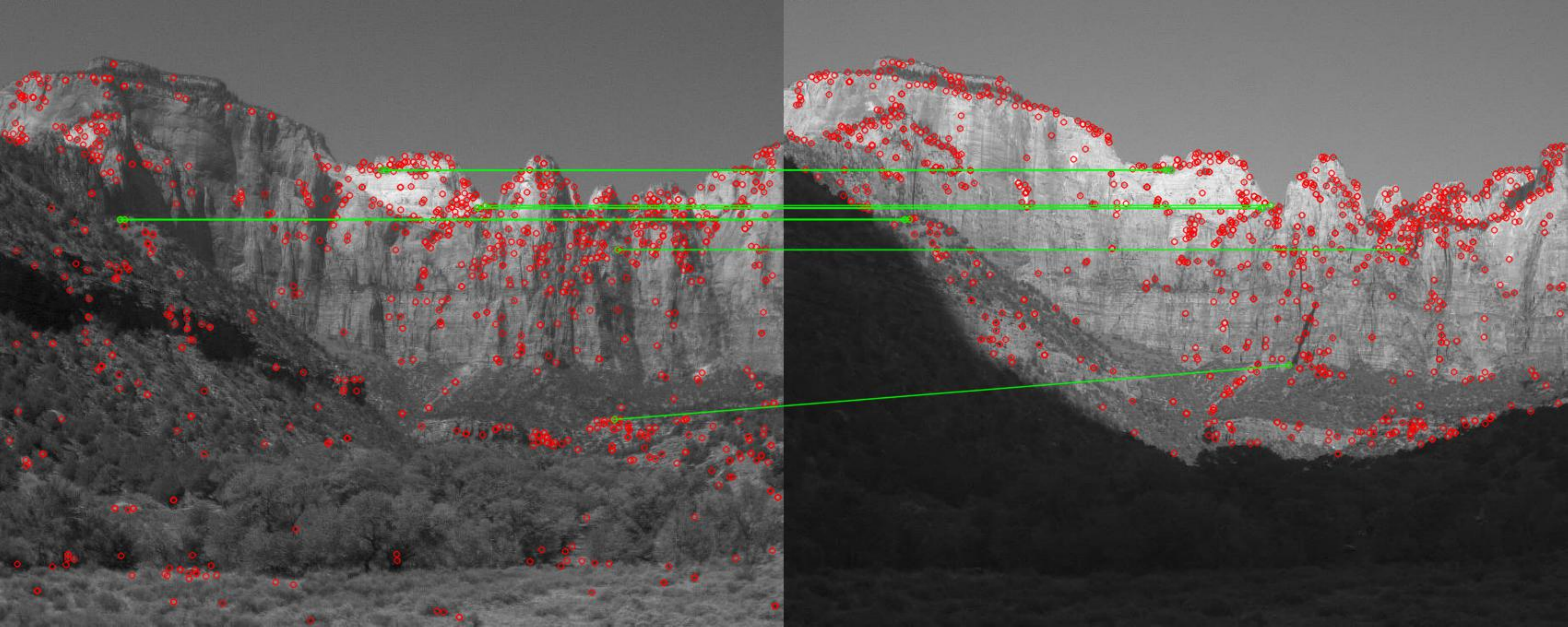} &
        \includegraphics[width=\sz\textwidth]{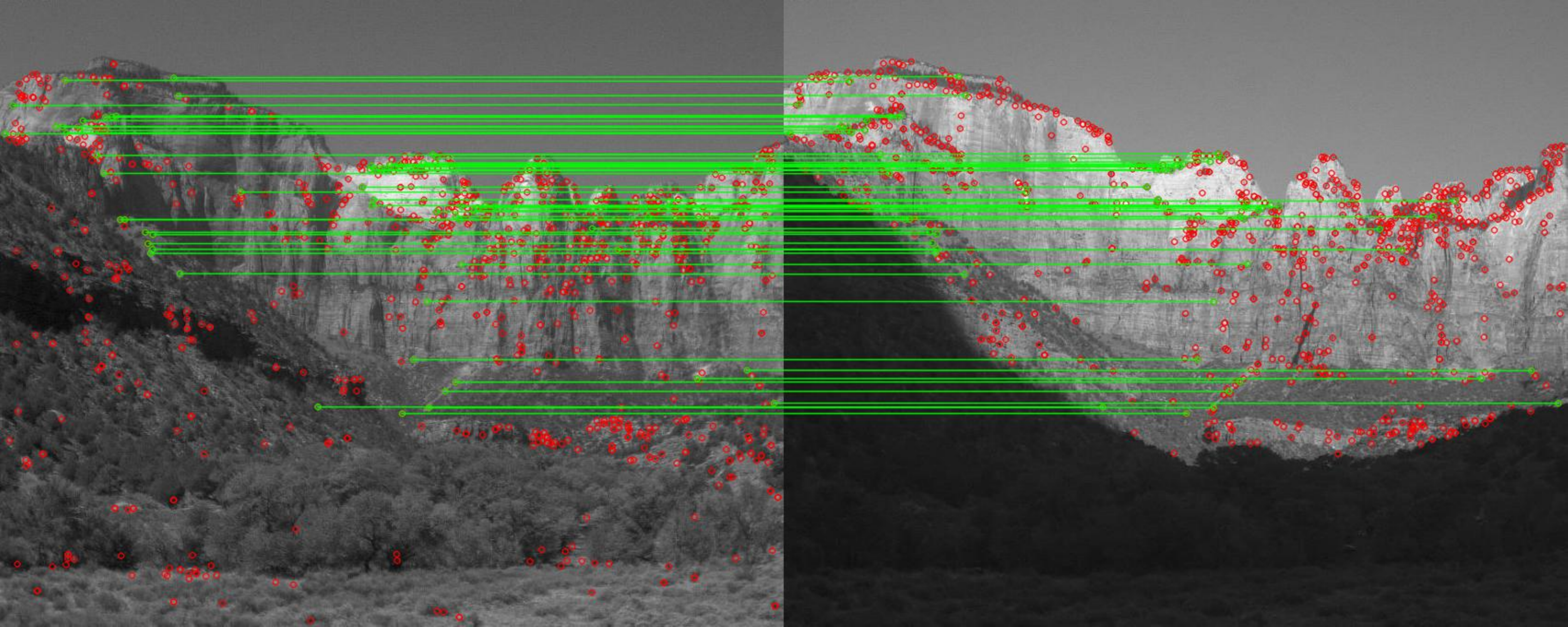} &
        \includegraphics[width=\sz\textwidth]{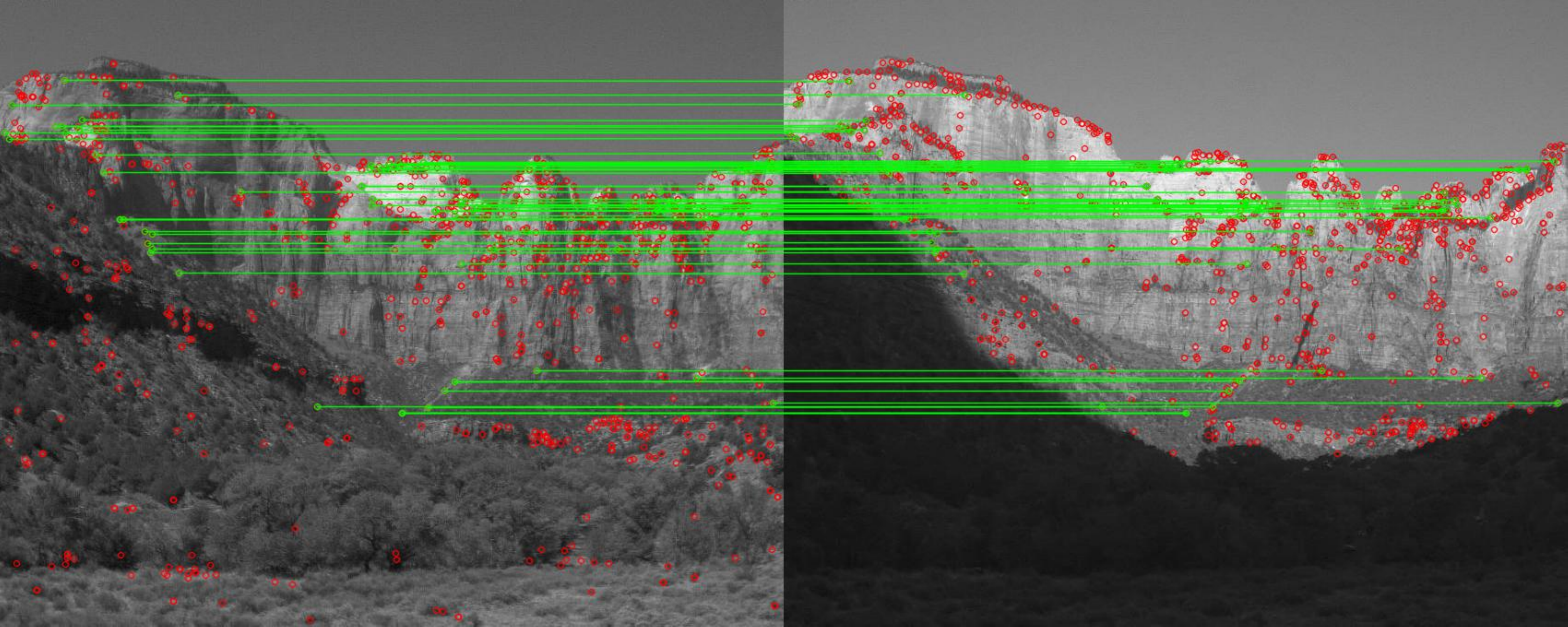}\\[-2mm]
    \end{tabular}
    \caption{\textbf{Importance of invariance among descriptors.} SIFT descriptors (left) perform well on rotated images (top), but are outperformed by Upright SIFT descriptors (middle) when no rotation is present (bottom). We propose a method (right) that automatically selects the proper invariance during matching time.}
    \label{fig:intro}
\end{figure}

While feature detectors should generally be invariant to be repeatable under different scenarios~\cite{Zhou2016W}, the same is not necessarily true for descriptors~\cite{Wu083dmodel}. There is a direct trade-off for descriptors between generalization and discriminative power. More invariance allows a better generalization, but produces descriptors that are less informative. Figure~\ref{fig:intro} shows that the rotation variant descriptor Upright SIFT performs better than its invariant counterpart SIFT when only small rotations are present in the data. We argue that the best level of invariance depends on the situation. As a consequence, this questions the recent trend of jointly learning detector and descriptor: they may have to be dissociated if one does not want the descriptor to be as invariant as the detector.

In this work we focus on learning descriptors only and propose to select at runtime the right invariance given the context. Instead of learning a single generic descriptor, we compute several descriptors with different levels of invariance. We then propose a method to automatically select the most suitable invariance during matching. We achieve this by leveraging the local descriptors to learn meta descriptors that can encode global information about the variations present in the image. At matching time, the local descriptors distances are weighted by the similarity of these meta descriptors to produce a single descriptor distance. Matches based on this distance can then be filtered using standard heuristics such as ratio test or mutual nearest neighbor.

Overall, our method, named Local Invariance Selection at Runtime for Descriptors (LISRD - pronounced as lizard), brings flexibility and interpretability into the feature description. When some image variations are known to be limited for a given application, one may directly use the most discriminative descriptor among all our learned local descriptors. However, it is usually hard to make such an assumption about the inter-image variations, and LISRD can instead automatically select the best invariance independently for each local region. Hence we are able to distinguish between different levels of variations within the same image (e.g. if half of the image is in the shadow but not the other half) and we show that this can improve the matching capabilities in comparison to using a single descriptor. The meta descriptors formulation is also not restricted to our proposed learned local descriptors, but can be easily generalized to most keypoint detectors and descriptors, as shown in Figure~\ref{fig:intro} where it is applied to SIFT and Upright SIFT. Furthermore, the meta description only adds a small overhead to the current pipelines of keypoint detection and description in terms of runtime and memory consumption, which makes it suitable for real time applications.
In summary, this work makes the following \textbf{contributions}:
\begin{itemize}[topsep=2pt,leftmargin=*]
    \item We show how to learn several local descriptors with multiple variance properties through a single network, in a similar spirit as in multi-task learning.
    \item We propose a light-weight meta descriptor approach to automatically select the best invariance of the local descriptors given the context.
    \item Our concept of meta descriptor and general approach of invariance selection can be easily transferred to most feature point detectors and descriptors, which we demonstrate for learned as well as traditional handcrafted descriptors. 
\end{itemize}

\section{Related work}

\boldparagraph{Learned local feature descriptors.}
The recent progress in deep learning has enabled learned local descriptors to outperform the classical baselines by a large margin~\cite{detone2018,Dusmanu2019CVPR,Luo2019,revaud2019}. Following the classical approach, early works run a CNN on a small image region around the point of interest to get a patch descriptor~\cite{yurun2017,mishchuk2017,ono2018lfnet}. The patch is not restricted to square areas, but can encode spatial transforms, such as affine~\cite{Mishkin_2018_ECCV} and polar~\cite{Ebel_2019_ICCV} ones. The network is often optimized with a triplet loss using heuristics to extract positive and negative patches~\cite{balntas2016pnnet,Luo2018,xufeng2015,sosnet2019cvpr}, or by directly maximizing the average precision (AP)~\cite{he2018}. Working on sparse features also gives the possibility to leverage both the visual context of the image and the spatial relationships between the keypoint locations~\cite{Luo2019}.
More recently, descriptors extracted densely by CNN architectures from full images have shown both fast inference time and high performance on matching and retrieval tasks, and can jointly detect a heatmap of keypoints. Some works detect keypoints and describe them in parallel, such as SuperPoint~\cite{detone2018} and R2D2~\cite{revaud2019}, with for the latter an additional reliability map keeping track of the most informative locations in the image. Another approach is to use the features of the network as dense descriptors to subsequently detect keypoints, based on those features~\cite{Noh_2017_ICCV,Dusmanu2019CVPR,yang2020ur2kid}. DELF~\cite{Noh_2017_ICCV} selects the keypoints using a learned attention, D2-Net~\cite{Dusmanu2019CVPR} retrieves the maximum responses of the descriptor feature map across all channels, while UR2KID~\cite{yang2020ur2kid} clusters the channels in different groups and extracts keypoints based on their L2 responses. 
Even though jointly estimating the keypoints and descriptors allows a faster prediction and yields descriptors that are more correlated to the keypoints, the consequence is that detector and descriptor will share the same invariance. Therefore, we choose to focus exclusively on descriptor learning in this work.

\boldparagraph{Invariance in feature descriptors.}
Selecting an online invariance for binary descriptors is the core idea of BOLD~\cite{balntas2015}, where a subset of the binary tests is chosen at runtime for each image patch to maximize the invariance to small affine transformations.
Similarly, the general trend of most recent learned methods is to obtain descriptors as invariant as possible to any image variations. 
LIFT~\cite{yi2016} mimics SIFT to achieve rotation invariance by estimating the keypoints, their orientation and finally their descriptor. 
Invariance to specific geometric changes can be achieved through group convolutions~\cite{cohen2016} by clustering the different geometrical transformations into specific groups~\cite{liu2019}. However, the usual strategy is to incorporate as much diversity in the training data as possible. Illumination invariance can for example be obtained by training on images with multiple lighting conditions~\cite{kaliroff2019}. Photometric and homographic data augmentations also increase robustness to illumination and viewpoint changes~\cite{detone2018}. Similarly, R2D2~\cite{revaud2019} improves the robustness to day-night changes by synthesizing night images with style transfer and also to viewpoint changes by leveraging flow between close-by images~\cite{revaud2015}. Methods like D2-Net~\cite{Dusmanu2019CVPR} and UR2KID~\cite{yang2020ur2kid} leverage a large database of images with multiple conditions and non planar viewpoint changes thanks to SfM data~\cite{MDLi18}. In this work, we adopt a mixture of the previously mentioned methods, namely the same synthesized night images as in~\cite{revaud2019}, homographic augmentation, and training on datasets with multiple illumination changes~\cite{murmann19}.

\boldparagraph{Multi-task learning in description and matching tasks.}
Using a single network to achieve multiple and related tasks in feature description and matching is not new. Jointly learning the detector and descriptor~\cite{detone2018,Dusmanu2019CVPR,revaud2019} is already multi-task learning that makes the descriptors more discriminative at the predicted keypoint locations. HF-Net~\cite{sarlin2019coarse} unifies the detection of feature points, local and also global descriptors for image retrieval using multi-task distillation with a teacher network. Methods such as SuperGlue~\cite{sarlin2019superglue} and ContextDesc~\cite{Luo2019} can leverage both visual and geometric context in their descriptors in order to get a more consistent matching between images. UR2KID~\cite{yang2020ur2kid} bypasses the need of keypoint supervision during training and directly optimizes the descriptors jointly for local matching and image retrieval. In our approach, multiple descriptors are also learned in parallel, but instead of differing in their scope, they differ in their level of invariance. 
Furthermore, unlike previous hierarchical global-to-local approaches, our method relies on local descriptors first and leverages global information only to refine the local matching.

\begin{figure}[tb]
    \centering
    \includegraphics[width=\textwidth]{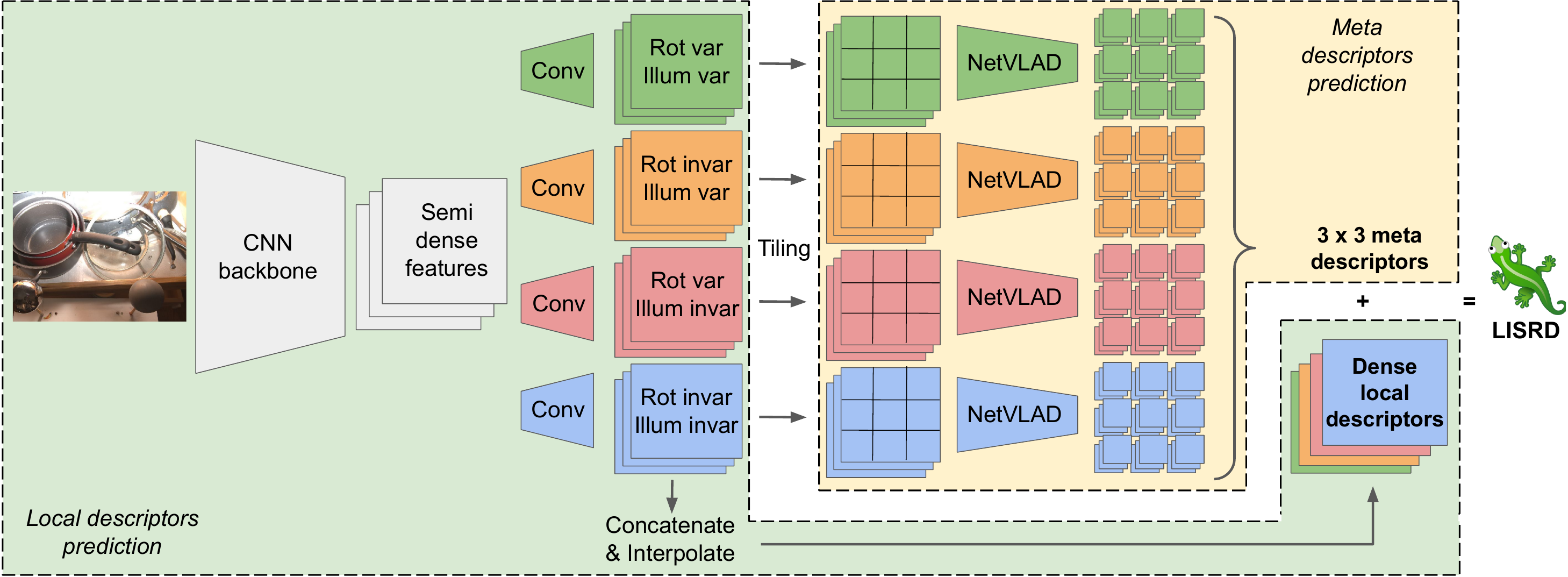}
    \caption{\textbf{Overview of our network architecture.} Our network computes four local dense descriptors with diverse invariances and aggregates them through a NetVLAD layer~\cite{Arandjelovic16} to obtain a regional description of the variations of the image. }
    \label{fig:overview}
\end{figure}

\section{Learning the best invariance for local descriptors}
%
Our approach to select the most relevant variance for local feature descriptors consists in two steps. 
First, we design a network to learn several dense descriptors, each with a different type of invariance (see Section~\ref{sec:local_desc}). 
Second, we propose a strategy in Section~\ref{sec:meta_desc} to determine the best invariance to use when matching the local descriptors. Figure~\ref{fig:overview} provides an overview of the full architecture.

\subsection{Disentangling invariance for local descriptors}
\label{sec:local_desc}
%
Many properties of an image have an influence on descriptors, but disentangling all of them would be intractable. We focus here on two factors known to have a large impact on descriptors performance: rotation and illumination. Our framework can however be generalized to other kinds of variations, for instance scaling. Since each of the two factors can either be variant or invariant, there are four possible combinations of variance with respect to illumination and rotation. We show in the following that the variant versions of descriptors are more discriminative since they are more specialized, while the invariant ones are trading the discriminative power for better generalization capabilities.

\boldparagraph{Network architecture.}
Our network is inspired by SuperPoint~\cite{detone2018}, with slight modifications. It takes RGB images as input, computes semi-dense features with a shared backbone of convolutions and is then divided into 4 heads predicting a semi-dense descriptor each, one per combination of variance, as shown in Figure~\ref{fig:overview}. Since most computations are redundant between the 4 local descriptors, the shared backbone reduces the number of weights in the network and offers an inference time competitive with the current learned descriptors.

\boldparagraph{Dataset preparation.}
The training dataset is composed of triplets of images. The first one, the \textit{anchor image} $I^A$, is taken from a large database of real images. The \textit{variant image} $I^V$ is a warped version of the anchor by a homography without rotation and with equal illumination to train variant descriptors. Finally, the \textit{invariant image} $I^I$ used for invariant descriptors is also related to the anchor by a homography, but its orientation and illumination can differ from the anchor.

\boldparagraph{Training losses.}
The local descriptors are trained using variants of the margin triplet ranking loss~\cite{balntas2016,mishchuk2017}, depending on whether the descriptor should be invariant or not to the variations present in $I^I$. The dense descriptors are first sampled on selected keypoints of the images, they are L2-normalized and the losses are computed on the resulting set of feature descriptors. Since we focus on descriptors only, we use SIFT keypoints during training to propagate the gradient in informative areas of the image only. Any kind of keypoint can be used at inference time nonetheless, as demonstrated in Section~\ref{sec:aachen_eval}. 

Formally, given two images $I^a$ and $I^b$ related by a homography $\mathcal{H}$ and $n$ keypoints $\mathbf{x}^a_{1..n}$ in image $I^a$, we warp each point to image $I^b$ using the homography: $\mathbf{x}^b_{1..n} = \mathcal{H}(\mathbf{x}^a_{1..n})$. This yields a set of $n$ correspondences between the two images, where we can extract the descriptors from each dense descriptor map: $\mathbf{d}^a_{1..n}$ and $\mathbf{d}^b_{1..n}$. Let us define a generic triplet loss $L_T(I^a, I^b, \mathrm{dist})$ between $I^a$ and $I^b$, given a descriptor distance $\mathrm{dist}(\mathbf{x}^a, \mathbf{x}^b)$. The triplet loss first enforces a correct correspondence $(\mathbf{x}_i^a, \mathbf{x}_i^b)$ to be close in descriptor space through a positive distance
\begin{equation}
    p_i = \mathrm{dist}(\mathbf{x}_i^a, \mathbf{x}_i^b) \ .
\end{equation}
Additionally, the triplet loss increases the negative distance $n_i$ between $\mathbf{x}_i^a$ and the closest point in $I^b$ which is at least at a distance $T$ from the correct match $\mathbf{x}_i^b$. This distance is computed symmetrically across the two images and the minimum is kept:
\begin{equation}
    n_i = \min(\mathrm{dist}(\mathbf{x}_i^a, \mathbf{x}_{n_b(i)}^b), \mathrm{dist}(\mathbf{x}_i^b, \mathbf{x}_{n_a(i)}^a)) \ ,
\end{equation}
with $n_b(i) = \argmin_{\scriptsize{j \in [1, n]}}(\mathrm{dist}(\mathbf{x}_i^a, \mathbf{x}_j^b)) \ \mathrm{s.t.} \ ||\mathbf{x}^a_i - \mathbf{x}^b_j||_2 > T$, and similarly for $n_a(i)$.
Given a margin $M$, the triplet margin loss is then defined as
\begin{equation}
    L_T(I^a, I^b, \mathrm{dist}) = \frac{1}{n} \sum_{i=1}^n \max(M + (p_i)^2 - (n_i)^2, 0) \ .
\end{equation}
In our case, the loss $L_I$ for invariant descriptors is an instance of this generic triplet loss between the anchor image $I^A$ and the invariant image $I^I$, for the L2 descriptor distance:
\begin{equation}
    L_I = L_T(I^A, I^I, ||\mathbf{d}^A - \mathbf{d}^I||_2) \ .
\end{equation}
The loss $L_V$ for variant descriptors is based on the full triplet of images: $I^A$, $I^I$ and $I^V$. It enforces variant descriptors to be different between the anchor and the invariant image, while preserving similarity between the anchor and the variant image. Its positive loss is the distance in descriptor space of positive matches between $I^A$ and $I^V$, and similarly for the negative distance between $I^A$ and $I^I$:
\begin{equation}
    L_V = \frac{1}{n} \sum_{i=1}^n \max(f M + ||\mathbf{d}^A_i - \mathbf{d}^V_i||_2^2 - ||\mathbf{d}^A_i - \mathbf{d}^I_i||_2^2, 0) \ ,
\end{equation}
where $f$ is a factor controlling at which point the anchor and the invariant images are different. For rotation changes, $f = \min(1, \frac{\theta_I}{\theta_{max}})$, where $\theta_I$ is the absolute angle of rotation between the anchor and the invariant image and $\theta_{max}$ is a hyper-parameter representing the threshold beyond which the two images should be considered different. This threshold ensures that only large rotations are penalized by the loss. It is hard to quantify the difference in illumination between two real images, so we set $f = 1$ when the illumination differs between the anchor and invariant image.

When a descriptor $d$ in the set $\mathcal{D}$ of descriptors is supposed to be invariant to all changes (illumination and/or rotation) between $I^A$ and $I^I$, we use $L_I$. Otherwise, $L_V$ is used. We define $L_{I/V}(d)$ as the selected loss and the total loss for local descriptors as
\begin{equation}
    L_l = \frac{1}{|\mathcal{D}|} \sum_{d\in\mathcal{D}} L_{I/V}(d) \ .
\end{equation}

\subsection{Online selection of the best invariance}
\label{sec:meta_desc}
%
\begin{figure}[tb]
    \centering
    \includegraphics[width=\textwidth]{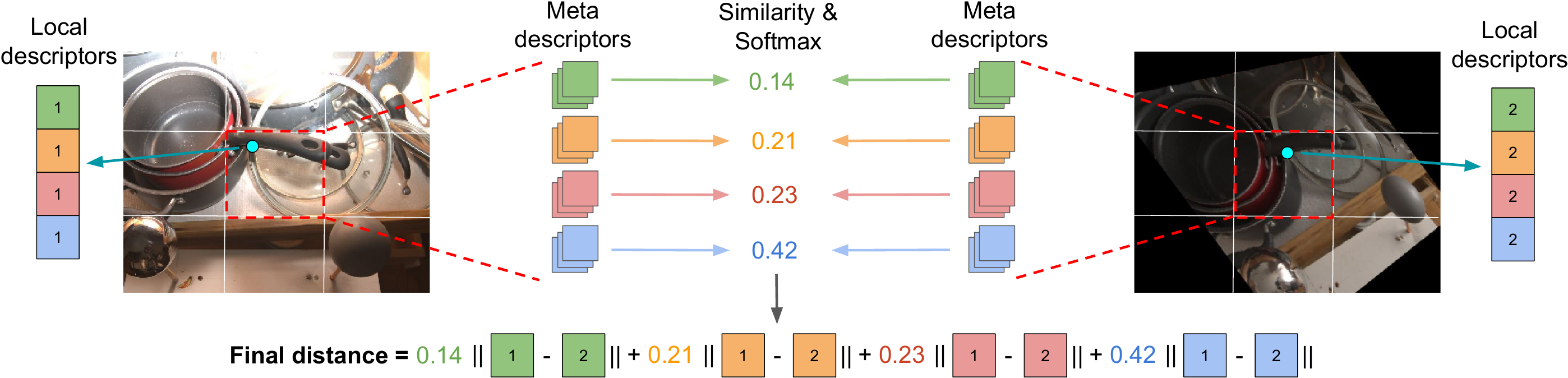}
    \caption{\textbf{The LISRD descriptor distance} between two points is the sum of the four local descriptors distances, weighted by the similarity of the meta descriptors.}
    \label{fig:distance_weighting}
\end{figure}

Given the local descriptors of the previous section, this section explores how to pick the most relevant invariance when matching images. Since it would be costly to recompute and compare the image variations for every pair of images to be matched, we propose to rely solely on the information contained in the descriptors to perform the selection. 
A naive approach would be to separately compute the similarity of the different local descriptors and to pick the most similar ones. However, the invariance selection would gain by having more context than the information of a single local descriptor and should be consistent with neighboring descriptors. Therefore, we propose to extract regional descriptors from the local ones and to use them to guide the invariance selection.

The local descriptors are thus gathered in neighboring areas through a NetVLAD layer~\cite{Arandjelovic16} to get a meta descriptor sharing the same kind of invariance as the subset of local descriptors, but with more context than a single local descriptor. Thus, having similar meta descriptors means sharing the same level of variations. The neighboring areas are created by tiling the image into a $c \times c$ grid and computing a meta descriptor for each tile. Hence, we get four meta descriptors per tile, which are then L2 normalized.

When matching the local descriptors of a tile, the four similarities between the meta descriptors are computed with a scalar product and we can rank the four local descriptors according to these similarities. Instead of making a hard choice by taking only the closest local descriptor, we use a soft assignment. A softmax operation is applied to the four similarities, to get four weights summing to one. These weights are then used to compute the distance between the local descriptors as shown in Figure~\ref{fig:distance_weighting}. 
More precisely, suppose that we want to compute the distance in descriptor space between point $\mathbf{x}^a$ in image $I^a$ and point $\mathbf{x}^b$ in image $I^b$. Point $\mathbf{x}^a$ is associated with 4 local descriptors $\mathbf{d}^a_{1..4}$ and 4 meta descriptors $\mathbf{m}^a_{1..4}$ corresponding to the region where $\mathbf{x}^a$ lies, and similarly for $\mathbf{x}^b$. Then the final descriptor distance between $\mathbf{x}^a$ and $\mathbf{x}^b$ is
\begin{equation}
    \mathrm{dist}(\mathbf{x}^a, \mathbf{x}^b) = \sum_{i=1}^4 \frac{\exp{((\mathbf{m}^a_i)^\intercal \cdot \mathbf{m}^b_i})}{\sum_{j=1}^4 \exp{((\mathbf{m}^a_j)^\intercal \cdot \mathbf{m}^b_j})} ||\mathbf{d}^a_i - \mathbf{d}^b_i||_2 \ .
\end{equation}
Thus, the similarity of the meta descriptors acts as a weighting of the local descriptors distances and can put a stronger emphasis on one specific variance when the corresponding meta descriptors have a high similarity. Matching is then performed with this descriptor distance, and can easily be refined with ratio test~\cite{lowe2004} or mutual nearest neighbor.

\boldparagraph{Training loss.}
The 4 NetVLAD layers are trained with a weak supervision based on another instance of the triplet loss $L_T$ between $I^A$ and $I^I$ with the distance defined above:
\begin{equation}
    L_m = L_T(I^A, I^I, \mathrm{dist})
\end{equation}
Thanks to this weak supervision, there is no need to explicitly supervise the meta descriptors, which would require knowing the amount of rotation and illumination for every tile in the image. The total loss of the network is finally a combination of the local and meta descriptors, weighted by a factor $\lambda$:
\begin{equation}
    L = L_l + \lambda L_m \ .
\end{equation}

\subsection{Training details}
%
\boldparagraph{Datasets.}
To train descriptors with different levels of variance in terms of rotation and illumination, datasets presenting all possible combinations of changes are needed. 
Control over the amount of changes is also required in order to know which loss between $L_I$ and $L_V$ should be used for each descriptor. We use in total four datasets to accomplish that. Illumination variations are obtained through the multi illumination dataset in the wild~\cite{murmann19} and the style transferred night images of the Aachen day dataset~\cite{revaud2019}. Both offer pairs of images with fixed viewpoint and different illuminations. Images with fixed illumination come from the MS COCO dataset~\cite{lin2014microsoft} and the day flow images from the Aachen dataset~\cite{revaud2019}. For all datasets except the latter, the images are augmented with random homographies containing translation, scaling, rotation and perspective distortion, similarly as in~\cite{detone2018}. For the day images of Aachen, the flow is used to create the correspondences and we consider that these images contain only small rotations and no major illumination changes. Overall, there is an equal distribution of images with and without illumination changes, and of rotated and non rotated images.

\boldparagraph{Implementation details.}
We describe here the details of our architecture. The backbone network, inspired by the VGG16~\cite{simonyan2014deep}, is composed of successive $3 \times 3$ convolutional layers with channel size 64-64-64-64-128-128-256-256. Each conv layer is followed by ReLU activation and batch normalization. Every two layers, a $2 \times 2$ average pooling with stride 2 is applied to reduce the spatial resolution by 2. For an image of size $H \times W \times 3$, the output feature map will have a size of $H/8 \times W/8 \times 256$. The local descriptor heads are all composed of the following operations: $3 \times 3$ conv of channel size 256 - ReLU - Batch Norm - $1 \times 1$ conv of channel size 128. The final dimension of each local descriptor is thus $H/8 \times W/8 \times 128$, and each concatenated descriptor is 512-dimensional. The semi-dense descriptors can then be bilinearly interpolated to the locations of any keypoint. Note that in order to achieve a better robustness to scale changes, one can also detect the keypoints and describe them at multiple image resolutions and aggregate the results in the original image resolution, similarly as in \cite{Dusmanu2019CVPR} and \cite{revaud2019}. The NetVLAD layers consists in 8 clusters of 128-dimensional descriptors, hence a meta descriptor size of 1024. We used $c \times c = 3 \times 3$ tiles per image.

The network is trained on RGB images resized to $240 \times 320$ with the following hyper-parameters: distance threshold $T = 8$, $\theta_{max} = \frac{\pi}{4}$, margin $M = 1$, loss factor $\lambda = 1$. It comprises roughly 3.7M parameters, which are optimized with the Adam solver~\cite{kingma2014} ($\mathrm{learning\ rate} = 0.001$ and $\beta = (0.9, 0.999)$). In practice, the local descriptors are pre-trained first and then fine-tuned by an end-to-end training with the meta descriptors. At test time, a single forward pass on a GeForce RTX 2080 Ti with $480 \times 640$ images takes 6ms on average.

\section{Experimental results}
%
We present here experiments validating the relevance of our method. Section~\ref{sec:method_validation} highlights the importance of learning different invariances, validates the proposed approach with an ablation study, and shows that LISRD can be extended to other descriptors such as SIFT and Upright SIFT. LISRD is then compared to the state of the art on a benchmark homography dataset (Section~\ref{sec:hp_eval}), on a challenging dataset with diverse conditions where the presence or lack of invariance is essential (Section~\ref{sec:dnim_eval}) and on a visual localization task in the real world (Section~\ref{sec:aachen_eval}).
%
\subsection{Metrics}
%
Since we want to compare the performance of the descriptors only, all the following metrics are computed on SIFT keypoints if not stated otherwise. The metrics are computed on pairs of images resized to $480 \times 640$ and related by a known homography. Resizing is performed by upscaling/downscaling the images to have each edge greater or equal respectively to $480$ and $640$, and a central crop is applied to get the target resolution. We keep a maximum of 1000 points among the keypoints shared between the two views and matches are obtained after mutual nearest neighbor filtering.

\boldparagraph{Homography estimation.}
We follow the procedure of \cite{detone2018} to compute a homography estimation score. Given a pair of images, RANSAC is used to fit a homography between the clouds of matched keypoints. The score is obtained by warping the four corners of the first image $\hat{c}_{1\ldots 4}$ with the predicted homography and comparing their distance to the same points $c_{1\ldots 4}$ warped by the ground truth homography. The homography is considered as correct when the average distance is below a threshold $\epsilon$, which is set to 3 pixels in all experiments: $\mathrm{HEstimation} = \frac{1}{4} \sum_{i=1}^4 ||\hat{c}_i - c_i||_2 \leq \epsilon$.

\boldparagraph{Precision.}
Precision (also known as mean matching accuracy) is the percentage of correct matches over all the predicted matches~\cite{Dusmanu2019CVPR,revaud2019}. We use by default a threshold of 3 pixels to consider a match to be correct.

\boldparagraph{Recall.}
Recall is the ratio of correctly predicted matches over the total number of ground truth matches, where a ground truth correspondence is the \emph{closest} point within an error threshold of 3 pixels. A predicted match with the \emph{second closest} point but still within the correct threshold is considered as incorrect.

\subsection{Method Validation}
\label{sec:method_validation}
\begin{figure}[tb]
    \centering
    \small
    \newcommand{\sz}{0.47}
    \begin{tabular}{cc}
        \hspace{2em}Illumination & \hspace{1.5em}Viewpoint\\[-0.5mm]
        \includegraphics[width=\sz\textwidth]{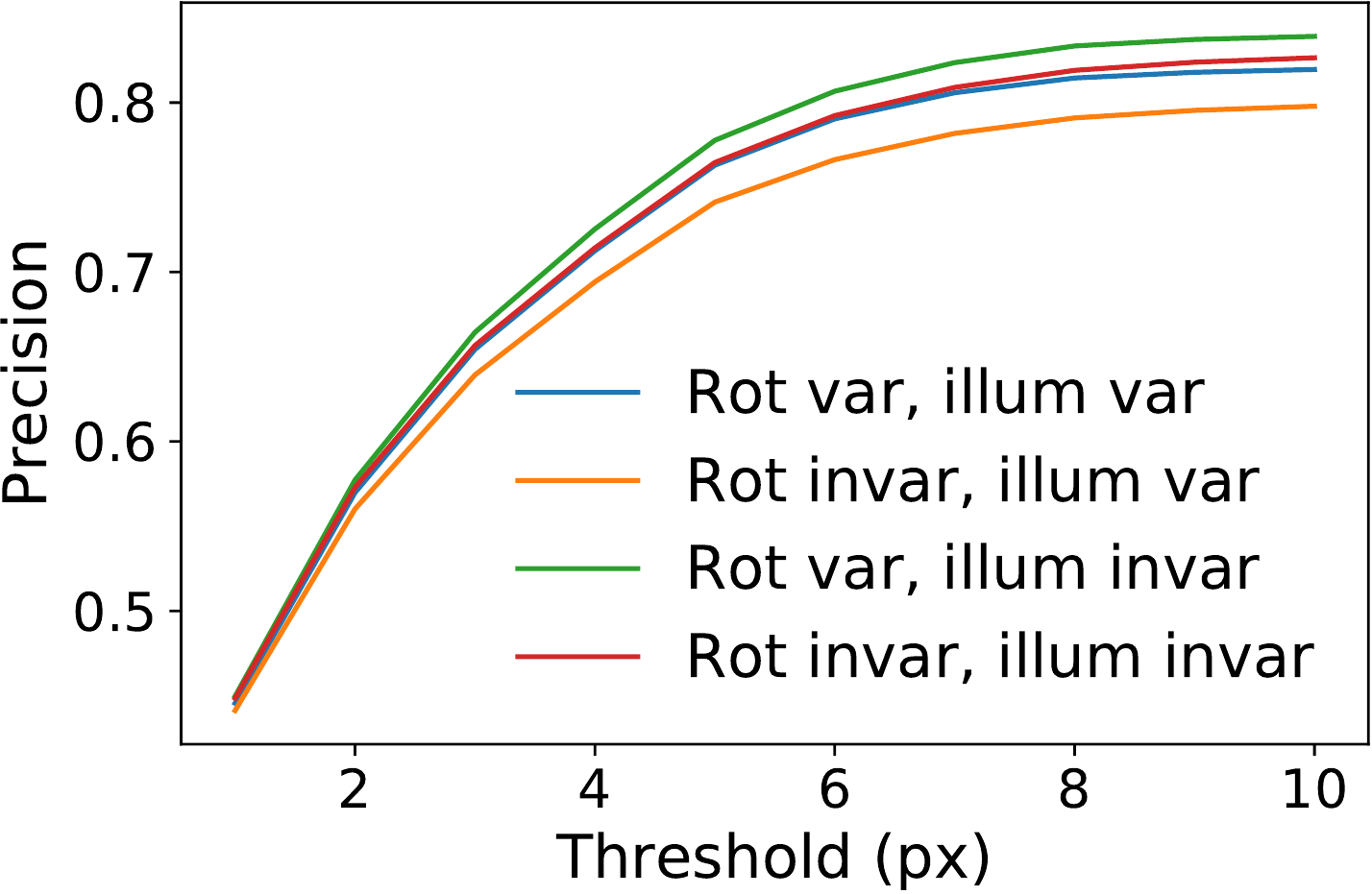} &
        \includegraphics[width=\sz\textwidth]{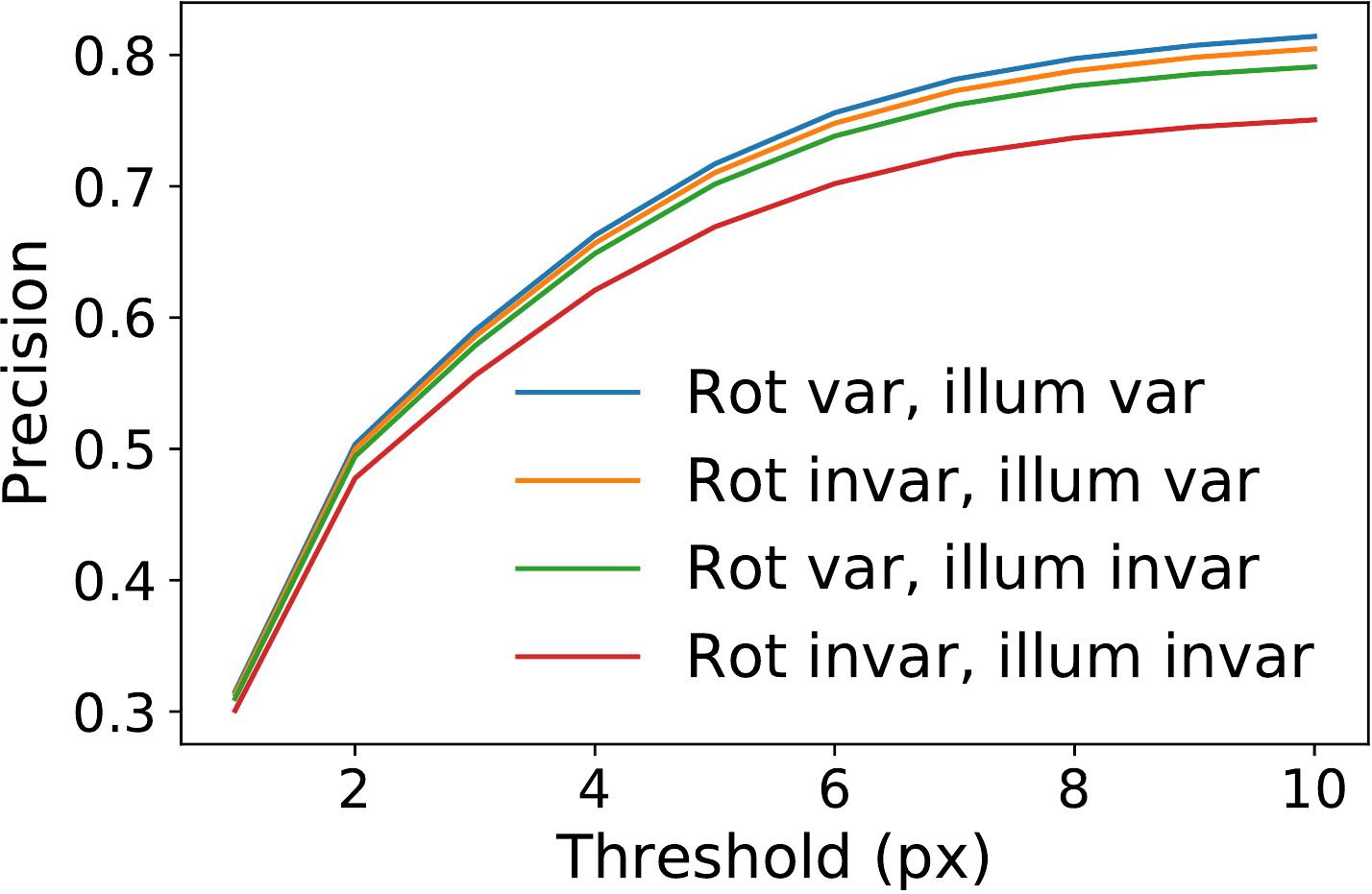} \\[-1em]
    \end{tabular}
    \caption{\textbf{Precision on HPatches of the 4 local descriptors.} Variant ones are better when invariance is not needed (e.g. rotation for the illumination dataset).}
    \label{fig:4descs_comparison}
\end{figure}

%
\boldparagraph{Impact of the different invariances.}
One can check the validity of our approach by comparing the 4 local descriptors. We use the HPatches dataset~\cite{hpatches_2017_cvpr}, which is standard in descriptor evaluation. It is composed of 116 sequences of 5 pairs of images, with either viewpoint changes (given by a known homography) or illumination changes with fixed viewpoint. Figure~\ref{fig:4descs_comparison} shows the comparison between the 4 descriptors in terms of precision. On viewpoint changes, the illumination variant descriptors are superior as the lighting is fixed in these images and they are thus more discriminative. Since HPatches contains few rotations, there is no significant difference in terms of rotation invariance and being rotation variant brings a small advantage on average. The precision on illumination changes shows that the best performing descriptors are the illumination invariant ones and that being rotation variant helps since the viewpoint is fixed. Thus there is no descriptor outperforming the others in all cases, and our hypothesis that variant descriptors are more discriminative than invariant ones is validated.

\boldparagraph{Ablation study.}
To confirm the benefit of our online selection of invariance and choice of parameters, we compare LISRD on homography estimation on the HPatches dataset with other selection methods of the local descriptors as well as with variants of our approach (Table~\ref{tab:ablation_study}). 
\textit{Best of the 4} computes the metrics for the 4 local descriptors separately and picks the best score. \textit{Greedy} computes the pairwise distances of all points for each local descriptor and greedily chooses the local descriptor with smallest distance for each pair of points. 
\textit{Hard assignment} selects the local descriptor that maximizes the meta descriptor similarity, instead of choosing a soft assignment as in our proposed method. 
\textit{No tiling} and $\mathit{5 \times 5}$ 
\textit{tiles} are variants of our method with no tiling or with $\mathit{5 \times 5}$ tiles for the meta descriptors. 
Finally, \textit{Single desc} is a descriptor trained with exactly the same architecture as ours, but with the 4 local descriptors concatenated and trained with invariance in both illumination and rotation.

On the full HPatches dataset, \textit{Best of the 4} corresponds to the descriptor invariant to both illumination and rotation, as both changes are present. However, our selection method can still leverage the other descriptors: for example an illumination variant descriptor for the viewpoint part. The disparity between LISRD and \textit{Greedy} and \textit{Hard assignment} highlights the added value of the meta descriptors, and shows that a soft assignment can better leverage the 4 descriptors at the same time. Finally, the comparison with \textit{Single desc} confirms our hypothesis that disentangling the types of invariance is beneficial compared to learning a single invariant descriptor with the same number of weights.

\begin{table}[tb]
    \begin{minipage}{0.4\textwidth}
        \centering
        \vspace{-2em}
        \captionof{table}{\textbf{Ablation study on the HPatches dataset.}}
        \label{tab:ablation_study}
        \scriptsize
        \setlength{\tabcolsep}{3mm}
        \begin{tabular}{lc}
            \toprule
             & HEstimation \\
            \midrule
            Best of the 4 & 0.778 \\[0.5mm]
            Greedy & 0.774 \\[0.5mm]
            Hard assignment & 0.762 \\[0.5mm]
            No tiling & 0.752 \\[0.5mm]
            $5 \times 5$ tiles & 0.773 \\[0.5mm]
            Single desc & 0.766 \\[0.5mm]
            LISRD (ours) & \textbf{0.784} \\
            \bottomrule
        \end{tabular}
    \end{minipage}
    \hspace{12pt}
    \begin{minipage}{0.55\textwidth}
        \centering
        \includegraphics[width=0.7\textwidth]{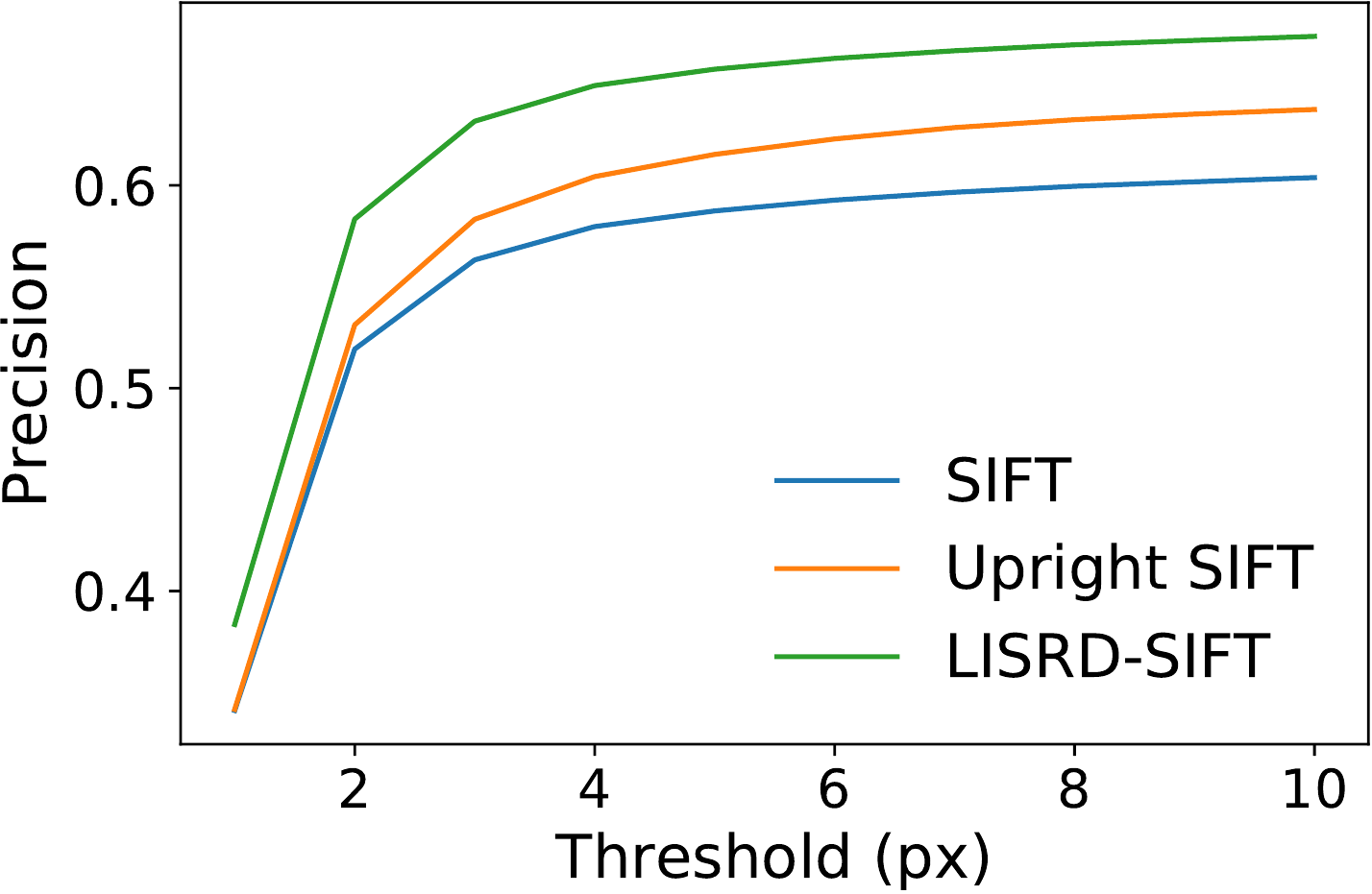}
        \captionof{figure}{\textbf{Variants of SIFT vs. our method fusing them (LISRD-SIFT).} Precision is computed on HPatches viewpoint.}
        \label{fig:fusion_sift_comparison}
    \end{minipage}
\end{table}

\boldparagraph{Generalization to other descriptors.}
LISRD can be easily generalized to other kinds of descriptors, and not only to our proposed learned local descriptors. We demonstrate this by applying our approach to the duo of local descriptors SIFT and Upright SIFT -- SIFT without rotation invariance, as presented in  Figure~\ref{fig:intro}. Instead of having four local descriptors, there are only two of them, one invariant to rotation and one variant, and similarly for the meta descriptors. Our method is evaluated against SIFT and Upright SIFT on the viewpoint part of HPatches. This dataset contains indeed sequences with no rotation, where Upright SIFT performs better, and other sequences with strong rotations, where SIFT takes over. Figure~\ref{fig:fusion_sift_comparison} shows that our method can effectively leverage both SIFT and Upright SIFT and outperforms the two.

\subsection{Descriptor evaluation on HPatches}
\label{sec:hp_eval}
%
This section compares the performance of LISRD against state-of-the-art local descriptors on the benchmark dataset HPatches. Since our approach requires global context from full images, we cannot run it on the patch level dataset. We use the full sequences of images instead, similarly as in~\cite{detone2018,Dusmanu2019CVPR,revaud2019}. We consider the following baselines: Root SIFT with the default Kornia~\cite{eriba2019kornia} implementation; HardNet~\cite{mishchuk2017} (trained on the PS-dataset~\cite{2018arXiv180101466M}), SOSNet~\cite{sosnet2019cvpr} (trained on the Liberty dataset of UBC Phototour~\cite{brown2010}), SuperPoint (SP)~\cite{detone2018}, D2-Net~\cite{Dusmanu2019CVPR}, R2D2~\cite{revaud2019} and GIFT~\cite{liu2019} with the authors implementation. Since we want to evaluate the descriptors only, SIFT keypoints are detected in the images and for each method, we extract the local descriptors at these locations. For Root SIFT, HardNet and SOSNet, we sample $32 \times 32$ patches at each SIFT keypoint and rotate them according to the SIFT orientation. As we want to evaluate the impact of rotation and illumination invariance only, we use single scale implementations for all methods\footnote{In the case of GIFT, which is both rotation and scale invariant, we sample images with scale 1 to make it rotation invariant only.}. Our method could however be made scale invariant using similar multi-scale approaches as in~\cite{Dusmanu2019CVPR,revaud2019}.

\begin{table}[tb]
    \centering
    \caption{\textbf{Comparison to the state of the art on HPatches.} Homography estimation, precision and recall are computed for error thresholds of 3 pixels. The best score is in bold and the second best one is underlined.}
    \scriptsize
    \setlength{\tabcolsep}{1.4mm}
    \begin{tabular}{llcccccccc}
        \toprule
         &  & Root SIFT & HardNet & SOSNet & SP & D2-Net & R2D2 & GIFT & Ours \\
        \midrule
        \multirow{3}{*}{\makecell{HP\\Illum}} & HEstimation & 0.898 & 0.884 & \underline{0.919} & 0.877 & 0.818 & 0.916 & \textbf{0.923} & 0.884 \\
         & Precision & 0.554 & 0.574 & 0.591 & 0.629 & 0.650 & \textbf{0.666} & 0.573 & \underline{0.665} \\
         & Recall & 0.431 & 0.483 & 0.519 & 0.565 & 0.564 & \underline{0.580} & 0.521 & \textbf{0.655} \\
        \midrule
        \multirow{3}{*}{\makecell{HP\\View}} & HEstimation & 0.644 & 0.688 & \textbf{0.742} & 0.651 & 0.553 & 0.627 & \underline{0.715} & 0.688 \\
         & Precision & 0.515 & 0.582 & \underline{0.598} & 0.595 & 0.564 & 0.550 & 0.552 & \textbf{0.626} \\
         & Recall & 0.350 & 0.422 & \underline{0.448} & 0.446 & 0.382 & 0.371 & 0.429 & \textbf{0.495} \\
        \bottomrule
    \end{tabular}
    \label{tab:hp_eval}
\end{table}

The results are summarized in Table~\ref{tab:hp_eval}. Overall, LISRD ranks among the two best methods in precision and recall. The possibility to leverage rotation variant descriptors on the fixed pairs of the illumination part and to alternatively select the right level of lighting invariance given the amount of illumination changes probably explains our superior performance on the illumination part. Note the comparison with SuperPoint, whose architecture and training procedure are very similar to LISRD, and where our method displays better results in all metrics, thus showing the gain of our approach. The weaker results in homography estimation can be explained by a limitation of our method. Since our meta descriptors have a very coarse spatial resolution ($3 \times 3$ grid), if one of them fails to pick the right invariance, this will impact all the matches of its region. Thus, the correct matches predicted by LISRD can in that case become very concentrated in a specific part of the image, which makes the homography estimation with RANSAC less accurate. This issue could be avoided with a finer tiling of the meta descriptors, but at the price of a reduced global context.

\subsection{Evaluation in challenging and cross-modal situations}
\label{sec:dnim_eval}
%
The HPatches dataset offers a fair benchmark, but is limited to only few rotations and medium illumination changes. Our approach is designed to be used in a variety of scenarios and with changing conditions, so that all our local descriptors can be leveraged. In order to evaluate our method on such a versatile task, we designed a new benchmark dataset, based on the day-night image matching (DNIM) dataset~\cite{Zhou2016W}. This dataset is composed of sequences of images of a fixed camera taking pictures at regular time intervals and across day and night, with a total of 1722 images. For each sequence, the image with timestamp closest to noon is taken as day reference and the image closest to midnight as night reference. 
We create two benchmarks, where the images of each sequence are paired with either the day reference or the night one. We then synthetically warp the pairs with the same homography sampling scheme as in~\cite{detone2018} with an equal distribution of homographies with and without rotations. We plan to release the homographies used in this benchmark to let other researchers compare with their own methods. Examples of images and matches for this dataset can be found in the supplementary material.

\begin{figure}[tb]
    \centering
    \includegraphics[width=\textwidth]{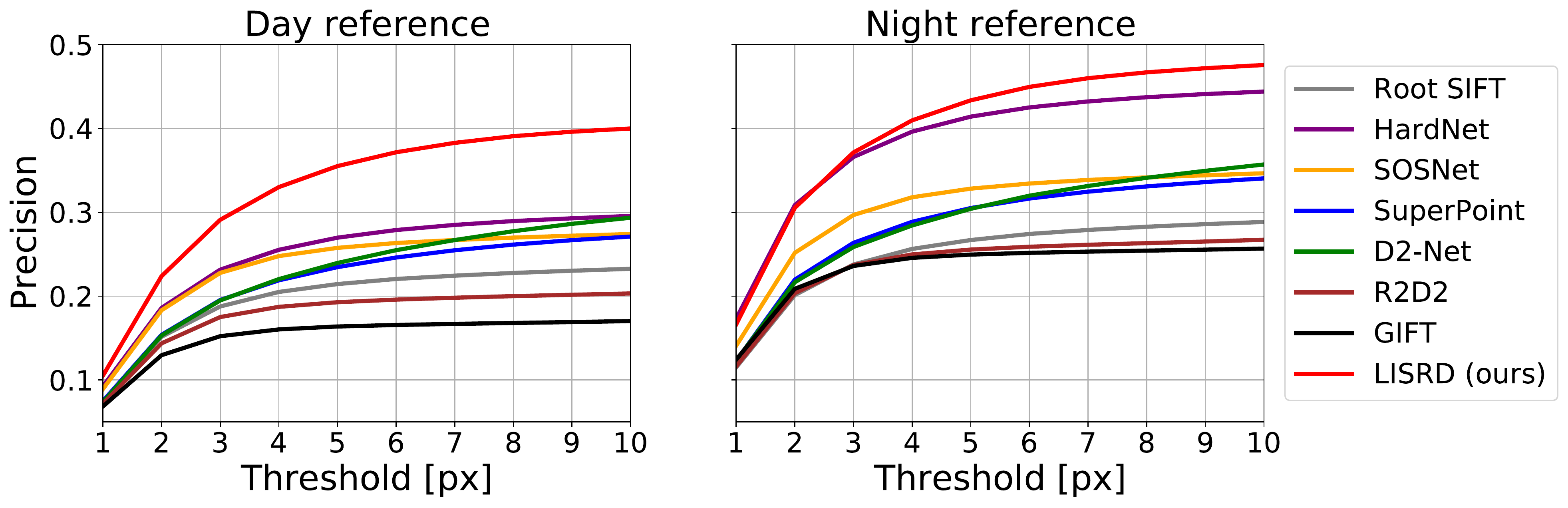}
    \vspace{-1.5em}
    \caption{\textbf{Precision curves on the DNIM dataset~\cite{Zhou2016W} augmented with rotations.} LISRD leverages its variant and more discriminative descriptors whenever possible and is thus more accurate than the state-of-the-art descriptors for all pixel error thresholds.}
    \label{fig:dnim_mma}
\end{figure}

Table~\ref{tab:dnim_eval} and Figure~\ref{fig:dnim_mma} show the evaluation with the state-of-the-art descriptors, using SuperPoint keypoints. LISRD can adapt its invariance to illumination and rotations to alternatively select the most relevant descriptor and it outperforms the other methods by a large margin both in terms of precision and recall.

\begin{table}[tb]
    \centering
    \caption{\textbf{Evaluation on a use case where invariance selection matters.} Homography estimation, precision and recall are computed with SuperPoint keypoints on a dataset with day-night changes and various levels of rotation. Selecting the relevant variant or invariant descriptors boosts the precision and recall of our method compared to the previous state-of-the-art methods.}
    \scriptsize
    \setlength{\tabcolsep}{1.4mm}
    \begin{tabular}{llcccccccc}
        \toprule
         &  & Root SIFT & HardNet & SOSNet & SP & D2-Net & R2D2 & GIFT & Ours \\
        \midrule
        \multirow{3}{*}{\makecell{Day\\ref}} & HEstimation & 0.121 & \textbf{0.199} & 0.178 & 0.146 & 0.094 & 0.170 & 0.187 & 0.198 \\
         & Precision & 0.188 & 0.232 & 0.228 & 0.195 & 0.195 & 0.175 & 0.152 & \textbf{0.291} \\
         & Recall & 0.112 & 0.194 & 0.203 & 0.178 & 0.117 & 0.162 & 0.133 & \textbf{0.317} \\
        \midrule
        \multirow{3}{*}{\makecell{Night\\ref}} & HEstimation & 0.141 & \textbf{0.262} & 0.211 & 0.182 & 0.145 & 0.196 & 0.241 & \textbf{0.262} \\
         & Precision & 0.238 & 0.366 & 0.297 & 0.264 & 0.259 & 0.237 & 0.236 & \textbf{0.371} \\
         & Recall & 0.164 & 0.323 & 0.269 & 0.255 & 0.182 & 0.216 & 0.209 & \textbf{0.384} \\
        \bottomrule
    \end{tabular}
    \label{tab:dnim_eval}
\end{table}

\subsection{Application to localization in challenging conditions}
\label{sec:aachen_eval}
%
A typical application of image matching including adverse conditions such as strong illumination changes and wide baselines is the visual localization task. We evaluate our method on the local feature challenge of CVPR 2019 based on the Aachen Day-Night dataset~\cite{Sattler_2018_CVPR}. The goal is to localize 98 night time query images as accurately as possible, given 20 day images per query with known camera pose. As the keypoint quality is essential in this task, we compare our method with other descriptors for various types of keypoints: SIFT, SuperPoint and D2-Net multi-scale (MS). 
The numbers for the baseline methods are taken from the benchmark on the official website\footnote{\url{https://www.visuallocalization.net/}}.
The results in Table~\ref{tab:aachen} show that our method is not limited to SIFT keypoints and can effectively improve the performance of local descriptors in challenging conditions. Note in particular the improvement over SuperPoint, which shares a similar architecture as ours.

\begin{table}[tb]
    \vspace{-1em}
    \centering
    \caption{\textbf{Visual localization performance on the Aachen Day-Night dataset~\cite{Sattler_2018_CVPR}.} We report the percentage of correctly localized queries for various distance and orientation error thresholds for SIFT, SuperPoint and D2-Net multi-scale (MS). Our method shows a good generalization when evaluated on different keypoints (KP) and can improve the original descriptor performance.}
    \scriptsize
    \setlength{\tabcolsep}{2.6mm}
    \begin{tabular}{ccccccc}
        \toprule
        \multirow{2}[2]{*}{\makecell{Error\\threshold}} & \multicolumn{2}{c}{SIFT KP} & \multicolumn{2}{c}{SuperPoint KP}  & \multicolumn{2}{c}{D2-Net KP}  \\
        \cmidrule(lr){2-3} \cmidrule(lr){4-5} \cmidrule(lr){6-7}
         & Up-Root SIFT                & Ours & SuperPoint                  & Ours & D2-Net (MS) & Ours (MS) \\
        \midrule
        $0.5m, 2^\circ$ & 54.1 & \textbf{72.4} & 73.5 & \textbf{78.6} & 67.3           & \textbf{73.5} \\
        $1m, 5^\circ$   & 66.3 & \textbf{82.7} & 79.6 & \textbf{86.7} & 87.8           & \textbf{88.8} \\
        $5m, 10^\circ$  & 75.5 & \textbf{94.9} & 88.8 & \textbf{98.0} & \textbf{100.0} & 99.0 \\
        \bottomrule
    \end{tabular}
    \label{tab:aachen}
\end{table}

\section{Conclusion}
%
We presented a novel approach to learn local feature descriptors able to adapt to multiple variations in images, while remaining discriminative. We unified the learning of several local descriptors with multiple levels of invariance  and of meta descriptors leveraging regional context to guide the local descriptors matching.

While restricted to illumination and rotation invariance, our framework can be generalized to more variations, at the cost of an exponentially growing number of descriptors however. A future direction of work would be to reduce the amount of redundancy between each descriptor by enforcing a stronger disentanglement separating each factors of variation. Since our approach is able to enforce different levels of invariance, one can also add another head to our network to predict invariant keypoints, while keeping discriminative descriptors, thus solving the current issue in joint learning of invariant detectors and descriptors.

Overall, this work is a first step towards disentangled descriptors. 
Separating the types of invariances paves the way to a full disentanglement of the factors of variations of images and could lead to flexible and interpretable local descriptors.

{\small
\vspace{0.2em}
\boldparagraph{Acknowledgments.}
This work has been supported by an ETH Zurich Postdoctoral Fellowship and Innosuisse funding (Grant No. 34475.1 IP-ICT).
\par
}

\clearpage
\bibliographystyle{splncs04}
\bibliography{paper}
\end{document}


\def\ECCVSubNumber{4158}
\title{Appendix}
\author{}
\institute{}
\maketitle

In the following, we provide additional qualitative and quantitative results about the Local Invariance Selection at Runtime for Descriptors (LISRD). Section~\ref{sec:rdnim} gives details about the benchmark dataset that was created to evaluate the impact of rotation and illumination invariance. Additional evaluations on multiple kinds of keypoint are available in Section~\ref{sec:other_kp}. Section~\ref{sec:dnim_timestamp} shows the performance of LISRD with respect to the state of the art for varying time intervals between the matched images. We also provide in Section~\ref{sec:aachen_day_night} an extended evaluation on the Aachen Day-Night dataset on the day split in addition to the night split. Finally, Section~\ref{sec:qualitative_examples} displays qualitative matches and selected invariances for challenging scenarios.

\section{A benchmark dataset for illumination and rotation invariances}
\label{sec:rdnim}
\begin{figure}[tb]
    \centering
    \newcommand{\sz}{0.48}
    \begin{tabular}{cc}
        \includegraphics[width=\sz\textwidth]{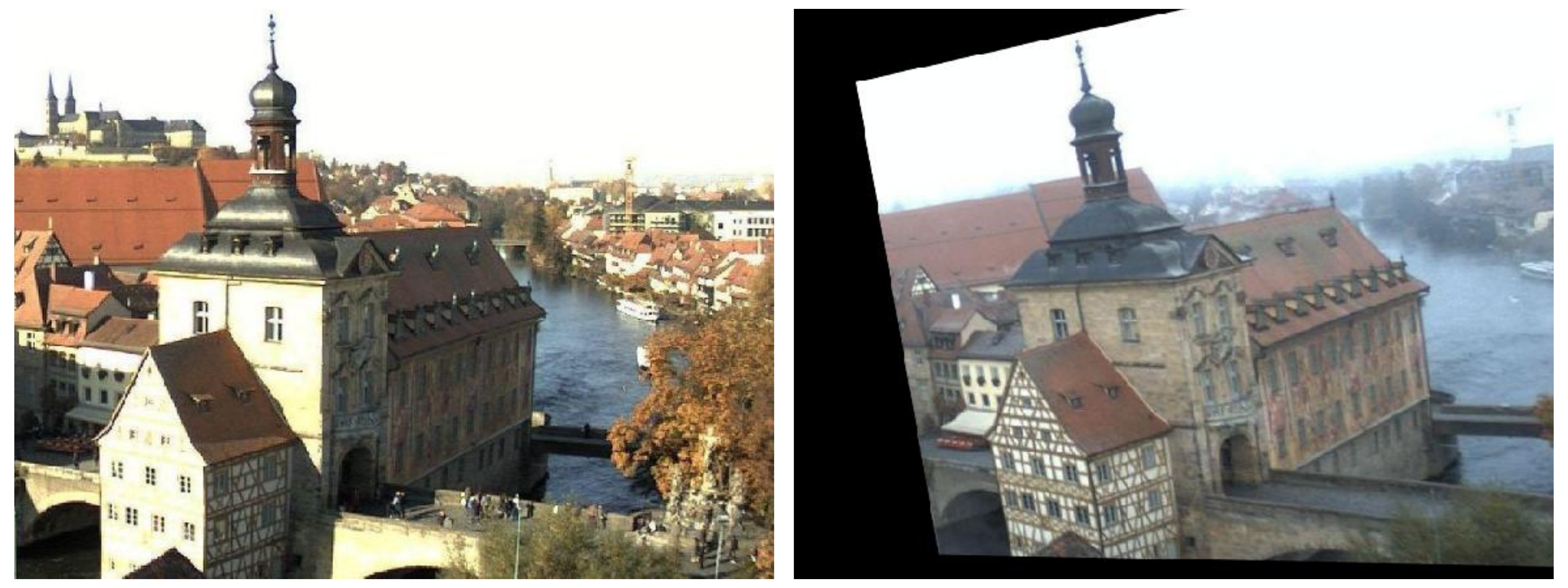}
        & \includegraphics[width=\sz\textwidth]{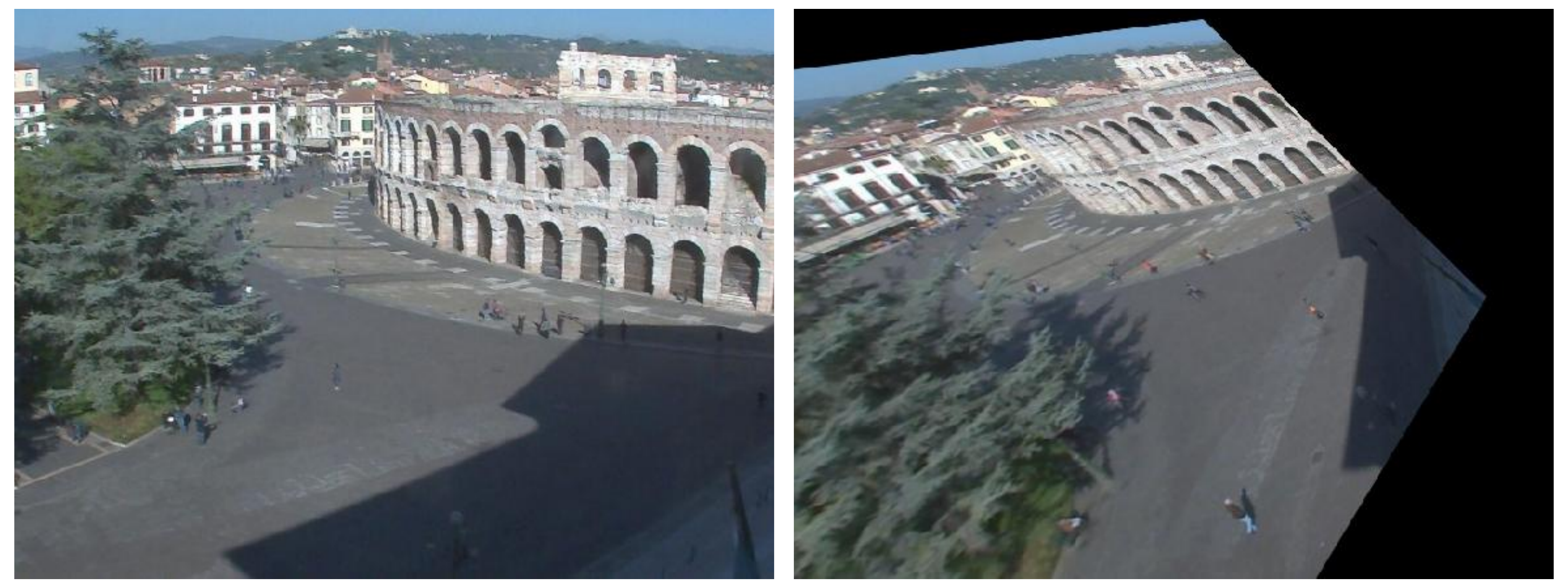} \\
        \includegraphics[width=\sz\textwidth]{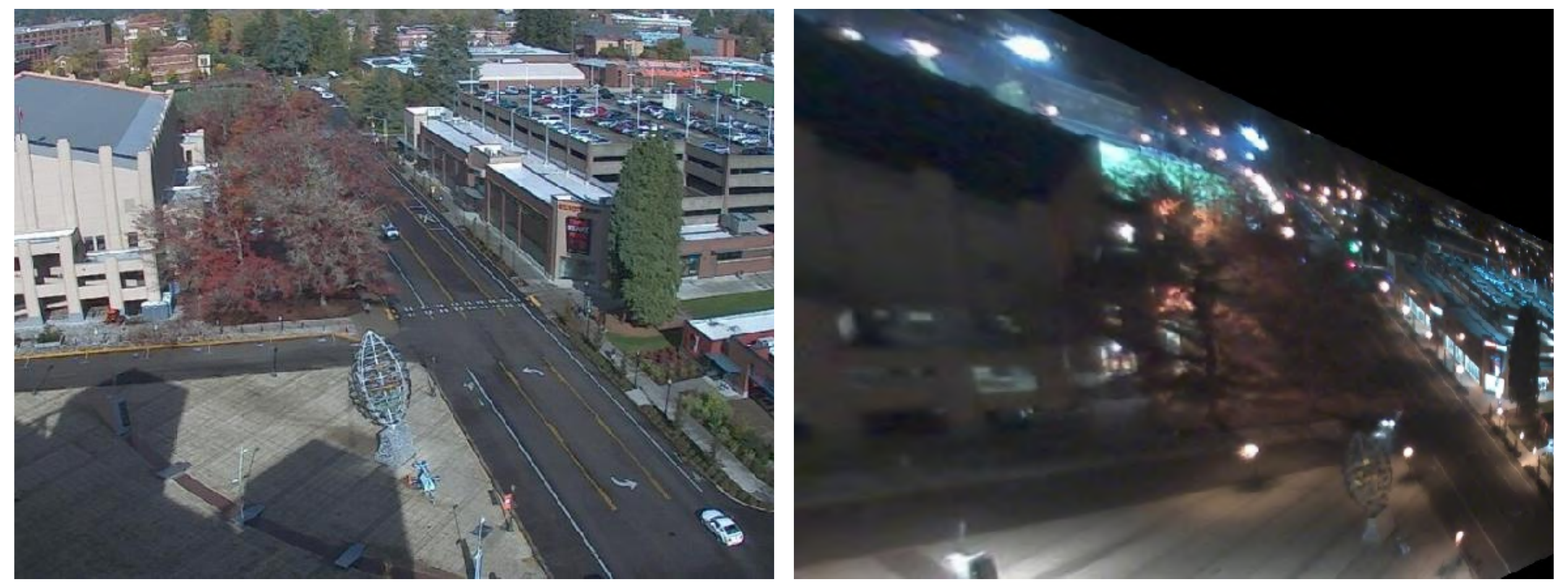}
        & \includegraphics[width=\sz\textwidth]{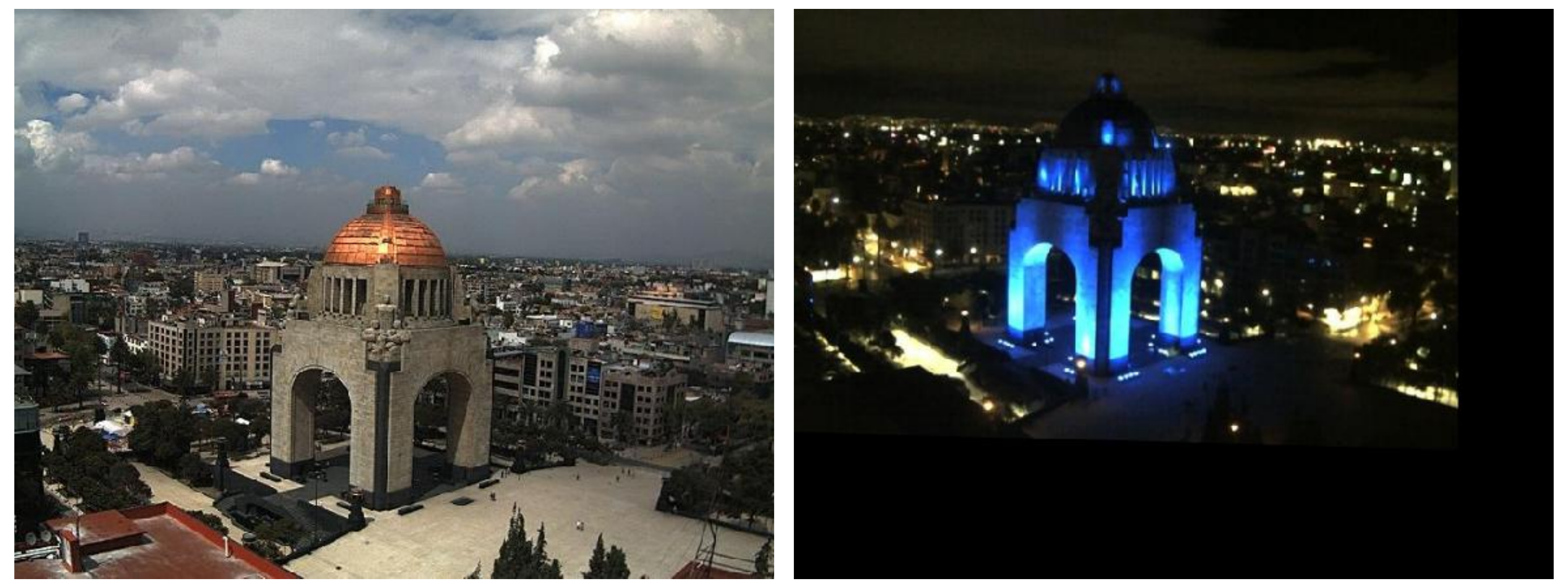} \\
        \includegraphics[width=\sz\textwidth]{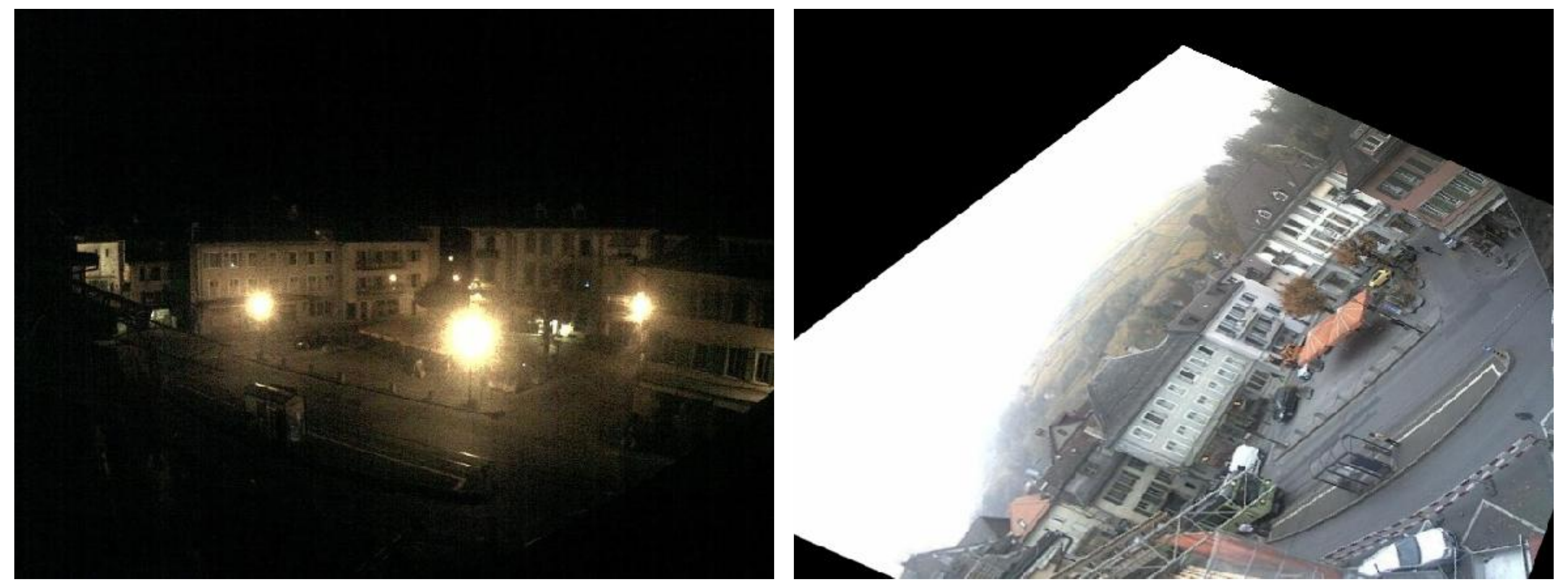}
        & \includegraphics[width=\sz\textwidth]{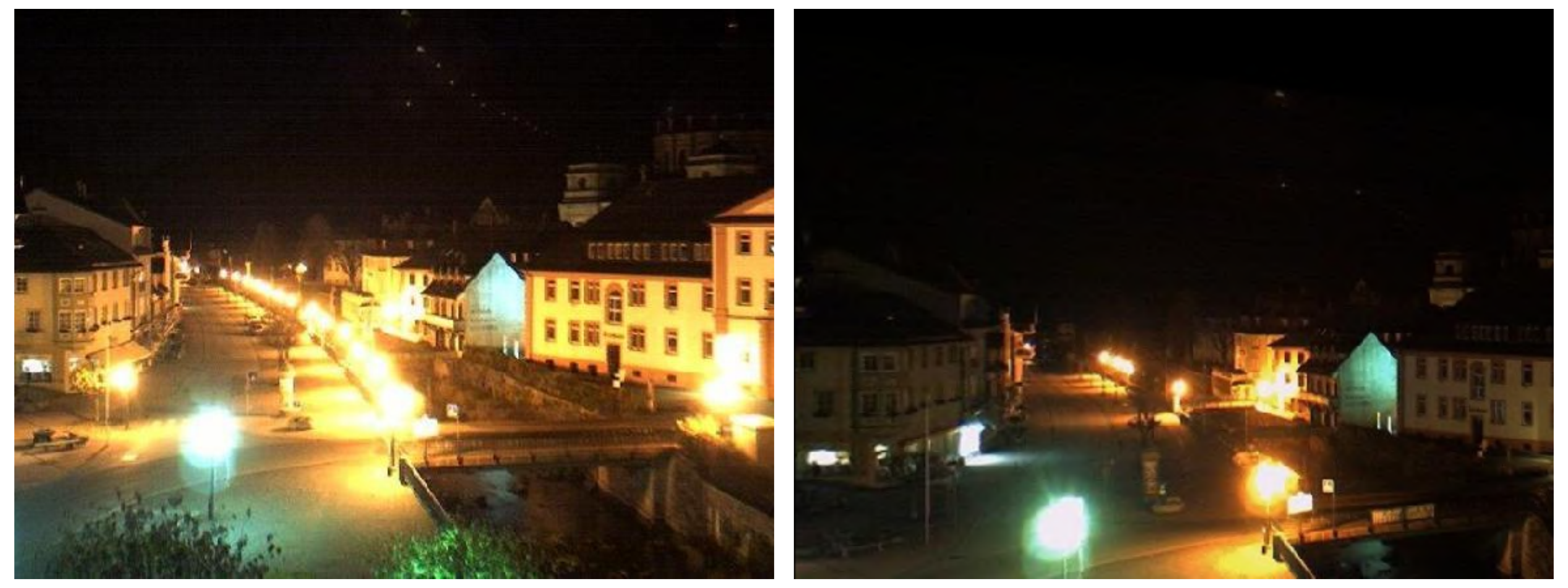}
    \end{tabular}
    \caption{\textbf{Sample images of the DNIM~\cite{Zhou2016W} dataset augmented with rotations.} All combinations among day-day, day-night and night-night pairs are available. Homographies are generated with random translations, rotations, scalings and perspective distortions, and images with and without rotation are equally distributed. When matching the images, the black artifacts created by the homography warping are masked out and ignored.}
    \label{fig:dnim_examples}
\end{figure}

The Day-Night Image Matching (DNIM)~\cite{Zhou2016W} dataset was originally released to evaluate the impact of day-night changes on local features matching. It consists of 1722 images grouped in 17 sequences of a fixed webcam taking pictures at regular time spans over 48h. In order to obtain pairs of images to match, a day and a night references are chosen for each sequence: the image with timestamp closest to noon is selected as day reference and similarly for the timestamp closest to midnight for the night reference. We then pair all the images in a sequence both with the corresponding day and night references, thus resulting in two benchmark datasets of 1722 pairs of images each. One dataset matches day references to all the DNIM images and is composed of day-day and day-night pairs, while the other dataset matches the night references to the DNIM images and displays night-night and night-day pairs. To simultaneously evaluate the robustness of our method to rotation and its discriminative power for non rotated images, we also warp the second image of each pair (i.e. the non reference image of the pair) with homographies. Similarly as in~\cite{detone2018}, these homographies are generated by combining random translations, rotations, scalings, and perspective distortions, with an equal distribution of rotated and non rotated images. Thus, we call this augmented dataset in the following RDNIM, for \emph{Rotated} DNIM. Examples of the RDNIM image pairs are available in Figure~\ref{fig:dnim_examples}.

\newpage

\section{Generalization to different keypoint detectors}
\label{sec:other_kp}
\begin{figure}[tb]
    \begin{subfigure}{\textwidth}
        \centering
        \includegraphics[width=\textwidth]{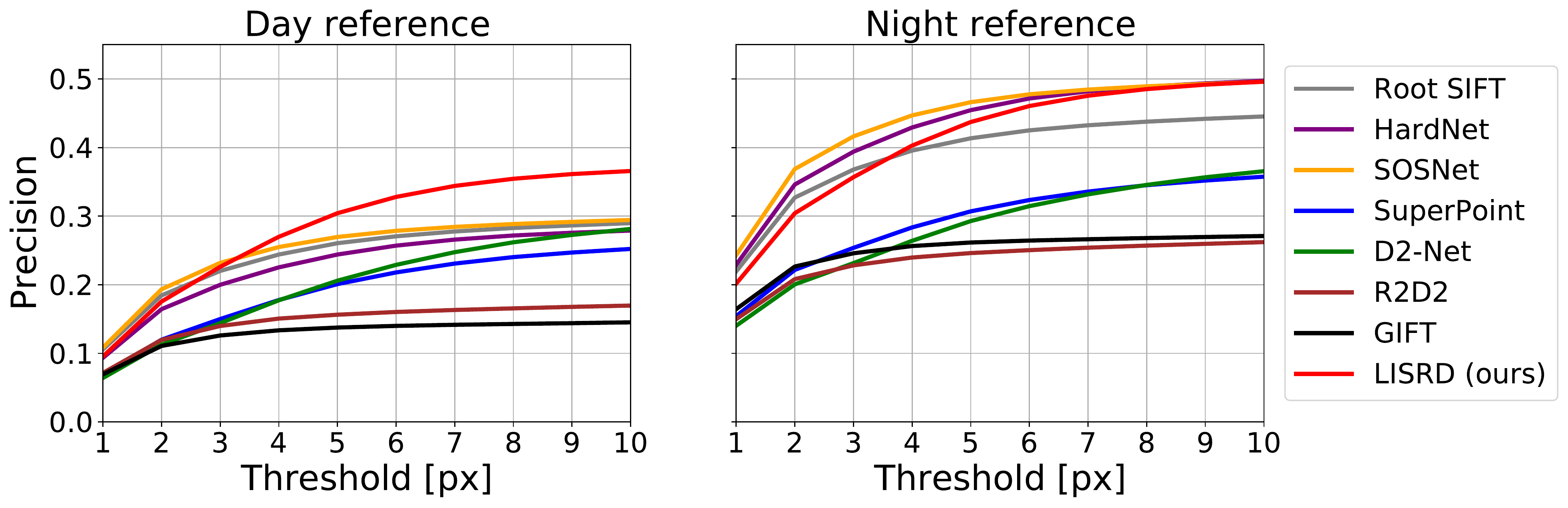}
        \caption{SIFT keypoints.}
        \label{fig:sift_kp}
        \vspace{1em}
    \end{subfigure}
    \vspace{-0.5em}
    \begin{subfigure}{\textwidth}
        \centering
        \includegraphics[width=\textwidth]{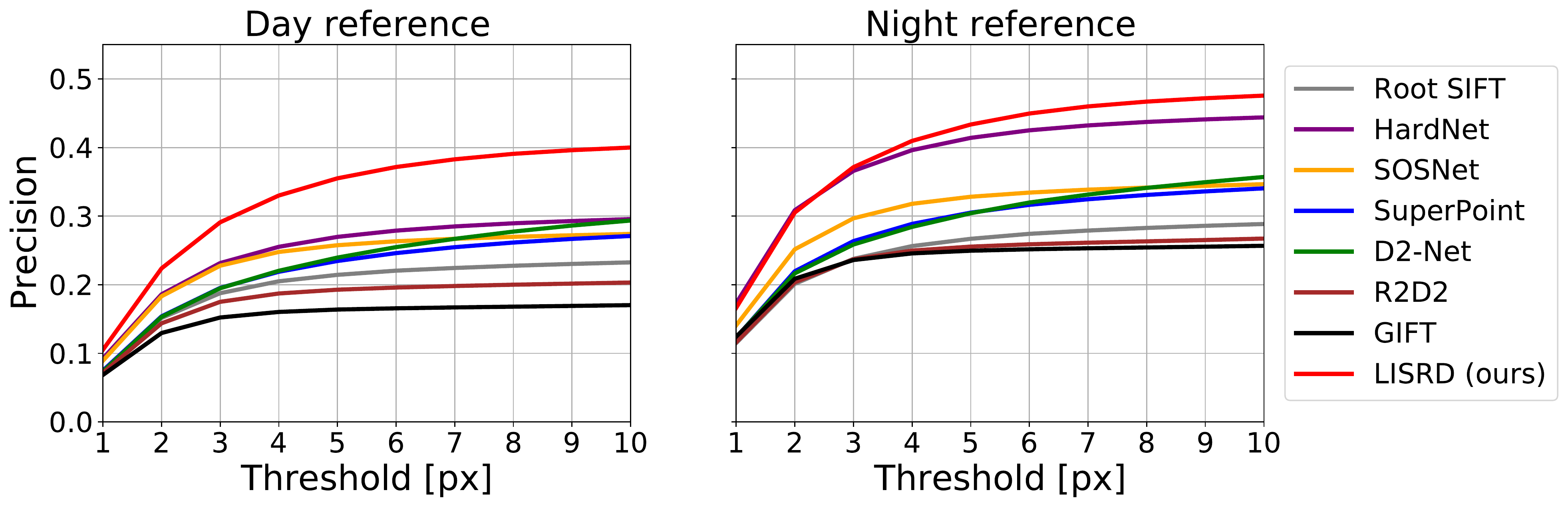}
        \caption{SuperPoint keypoints.}
        \label{fig:sp_kp}
        \vspace{1em}
    \end{subfigure}
    \vspace{-0.5em}
    \begin{subfigure}{\textwidth}
        \centering
        \includegraphics[width=\textwidth]{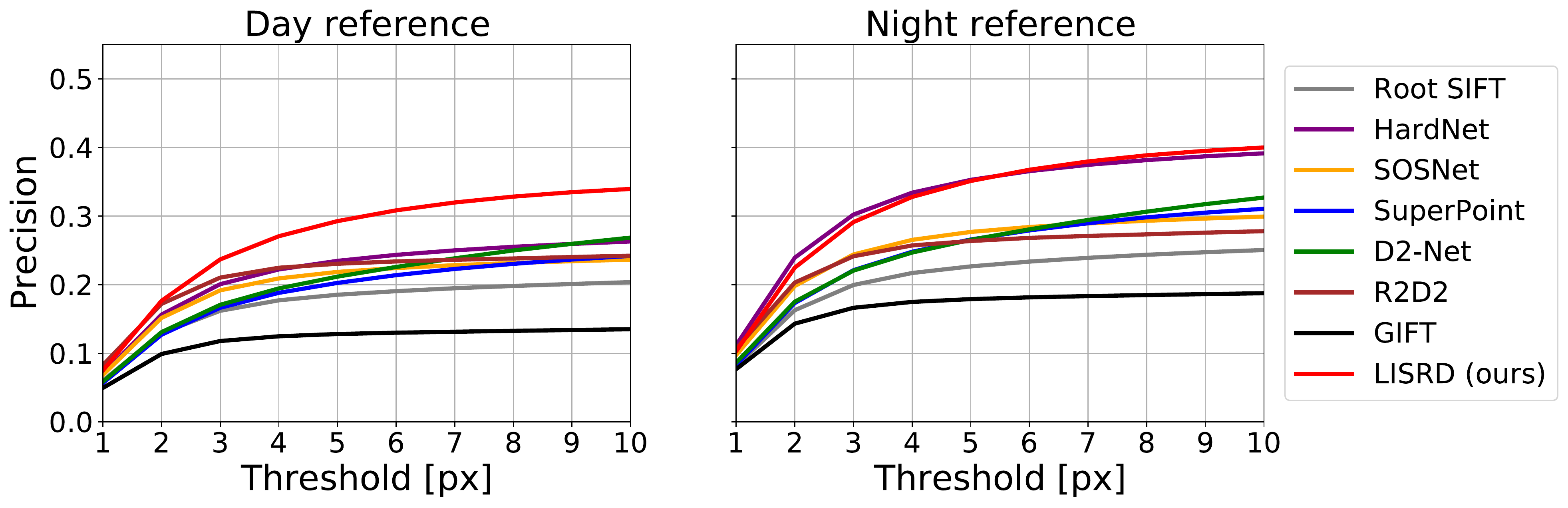}
        \caption{R2D2 keypoints.}
        \label{fig:r2d2_kp}
    \end{subfigure}
    \vspace{-0.5em}
    \caption{\textbf{Precision curves with SIFT~\cite{lowe2004}, SuperPoint~\cite{detone2018} and R2D2~\cite{revaud2019} keypoints on the RDNIM dataset.} The discriminative power of LISRD descriptors is not limited to SIFT keypoint locations, but also shows a high precision compared to the state of the art on other keypoints.}
    \label{fig:dnim_mma}
\end{figure}

LISRD was trained using SIFT keypoints~\cite{lowe2004}, but it can be used at test time with any other keypoints. We demonstrate this by providing additional comparisons to the state of the art on the RDNIM dataset, using SIFT, SuperPoint~\cite{detone2018} and R2D2~\cite{revaud2019} keypoints. Table~\ref{tab:dnim_eval} presents the evaluation with homography estimation, precision and recall for an error threshold of 3 pixels and Figure~\ref{fig:dnim_mma} shows the precision curves at multiple error thresholds. Overall, LISRD is competitive with the state of the art (HardNet and SOSNet) on SIFT keypoints and ranks first with learned keypoints on most metrics. Note the improved homography estimation score with learned keypoints, probably because these keypoints are well spread across the image and the limitation of LISRD mentioned in the main paper is curtailed. Indeed, this limitation was due to RANSAC producing bad estimates when the invariance selection failed in some regions and all the matches became concentrated in a small area. This phenomenon is less likely to happen when the keypoints are covering the whole image, and LISRD is thus able to get a more accurate homography estimation. Note that for each keypoint, the associated descriptor is not necessarily performing better, except for R2D2 that gets a slight improvement in precision when evaluated on their own keypoints. This is due to the reliability map used during their training, which makes their descriptors more discriminative at their keypoint locations.

As a feature direction of work, LISRD would benefit from learning its own keypoints with an additional head. This single head would predict invariant keypoints trained on images with multiple lightings and rotations and could be used with all descriptors - whether they are variant or not. This would ensure a better correlation between the keypoints and their descriptors and offer a faster prediction, instead of predicting separately keypoints and descriptors as is currently the case.

\begin{table}[tb]
    \centering
    \caption{\textbf{Evaluation with SIFT~\cite{lowe2004}, SuperPoint (SP)~\cite{detone2018} and R2D2~\cite{revaud2019} keypoints on the RDNIM dataset.} Homography estimation, precision and recall are computed for an error threshold of 3 pixels. LISRD is not restricted to the SIFT keypoints that were used during its training, but can be generalized to any keypoints (KP). The best score is in bold and the second best one is underlined.}
    \scriptsize
    \setlength{\tabcolsep}{1.2mm}
    \begin{tabular}{lllcccccccc}
        \toprule
         & &  & \makecell{Root\\SIFT} & HardNet & SOSNet & SP & D2-Net & R2D2 & GIFT & \makecell{LISRD\\(Ours)} \\
        \midrule
        \multirow{6}{*}{\makecell{SIFT\\KP}} & \multirow{3}{*}{\makecell{Day\\ref}} & HEstimation & 0.166 & \underline{0.170} & \textbf{0.215} & 0.084 & 0.057 & 0.121 & 0.145 & 0.127 \\
        & & Precision & 0.220 & 0.200 & \textbf{0.232} & 0.150 & 0.144 & 0.140 & 0.126 & \underline{0.226} \\
        & & Recall & 0.113 & 0.155 & \underline{0.197} & 0.114 & 0.081 & 0.107 & 0.108 & \textbf{0.212} \\
        \cmidrule{2-11}
        & \multirow{3}{*}{\makecell{Night\\ref}} & HEstimation & 0.255 & \underline{0.278} & \textbf{0.307} & 0.156 & 0.118 & 0.167 & 0.215 & 0.204 \\
        & & Precision & 0.368 & \underline{0.394} & \textbf{0.416} & 0.254 & 0.231 & 0.228 & 0.246 & 0.357 \\
        & & Recall & 0.212 & \underline{0.288} & \textbf{0.316} & 0.183 & 0.135 & 0.162 & 0.183 & 0.284 \\
        \midrule
        \multirow{6}{*}{\makecell{SP\\KP}} & \multirow{3}{*}{\makecell{Day\\ref}} & HEstimation & 0.121 & \textbf{0.199} & 0.178 & 0.146 & 0.094 & 0.170 & 0.187 & \underline{0.198} \\
        & & Precision & 0.188 & \underline{0.232} & 0.228 & 0.195 & 0.195 & 0.175 & 0.152 & \textbf{0.291} \\
        & & Recall & 0.112 & 0.194 & \underline{0.203} & 0.178 & 0.117 & 0.162 & 0.133 & \textbf{0.317} \\
        \cmidrule{2-11}
        & \multirow{3}{*}{\makecell{Night\\ref}} & HEstimation & 0.141 & \textbf{0.262} & 0.211 & 0.182 & 0.145 & 0.196 & \underline{0.241} & \textbf{0.262} \\
        & & Precision & 0.238 & \underline{0.366} & 0.297 & 0.264 & 0.259 & 0.237 & 0.236 & \textbf{0.371} \\
        & & Recall & 0.164 & \underline{0.323} & 0.269 & 0.255 & 0.182 & 0.216 & 0.209 & \textbf{0.384} \\
        \midrule
        \multirow{6}{*}{\makecell{R2D2\\KP}} & \multirow{3}{*}{\makecell{Day\\ref}} & HEstimation & 0.107 & \underline{0.187} & 0.181 & 0.140 & 0.093 & 0.135 & 0.157 & \textbf{0.193} \\
        & & Precision & 0.162 & 0.201 & 0.192 & 0.166 & 0.171 & \underline{0.210} & 0.118 & \textbf{0.237} \\
        & & Recall & 0.093 & 0.167 & \underline{0.172} & 0.168 & 0.101 & 0.076 & 0.102 & \textbf{0.290} \\
        \cmidrule{2-11}
        & \multirow{3}{*}{\makecell{Night\\ref}} & HEstimation & 0.135 & \textbf{0.196} & 0.168 & 0.145 & 0.101 & 0.132 & 0.183 & \underline{0.189} \\
        & & Precision & 0.200 & \textbf{0.302} & 0.244 & 0.221 & 0.221 & 0.241 & 0.166 & \underline{0.291} \\
        & & Recall & 0.132 & \underline{0.260} & 0.215 & 0.230 & 0.149 & 0.110 & 0.147 & \textbf{0.335} \\
        \bottomrule
    \end{tabular}
    \label{tab:dnim_eval}
\end{table}

\section{Evaluation across a full day}
\label{sec:dnim_timestamp}
\begin{figure}
    \centering
    \includegraphics[width=\textwidth]{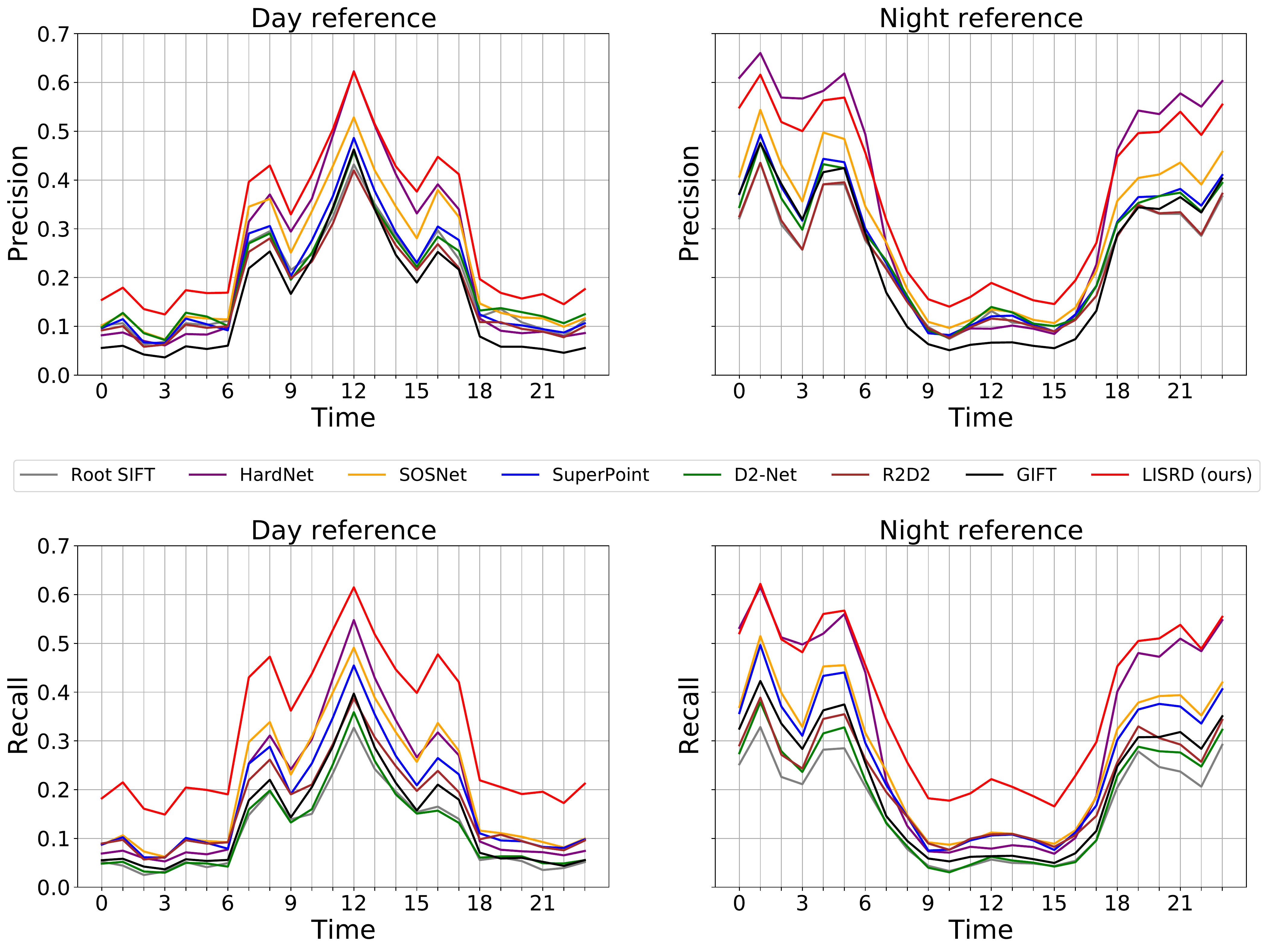}
    \caption{\textbf{Precision and recall across the day.} Precision and recall are computed on the RDNIM dataset with SuperPoint keypoints and an error threshold of 3 pixels. They are averaged for each hour of the day, based on the timestamp of the second image. The performance gradually degrades when the timestamp of the second image moves away from the reference time (noon for the day reference and midnight for the night reference). For close timestamps, LISRD leverages its illumination variant descriptors, but switches to the invariant ones when the timestamps differ too much. Thus, LISRD remains competitive with state-of-the-art descriptors for close timestamps and outperforms them when significant illumination changes are present.}
    \label{fig:dnim_over_time}
\end{figure}

The evaluation on the RDNIM dataset shows the global performance across a mix of day-day and day-night, or night-night and night-day images. But it is also interesting to study the performance at various times during the day. Figure~\ref{fig:dnim_over_time} displays the precision and recall curves on the RDNIM dataset along a full day. For every image in the second pair, we extract the hour at which the picture was taken from the timestamp, and round it to the closest integer. For each hour, the precision and recall are then computed and averaged across all images corresponding to this time and these averaged numbers are then plotted over the twenty-four hours of a day. We naturally get two peak curves, one centered at noon for the pairs with the day reference and the other centered at midnight for the night reference. LISRD is overall better than the other descriptors and, interestingly, the largest improvements come from the time intervals with day-night illumination changes. Thus, LISRD leverages its illumination variant and more discriminative descriptors when the timestamp of both images of the pair are close, and it switches to the invariant and more general ones when the images are taken at different times of the day.

\section{Evaluation on Aachen Day-Night dataset}
\label{sec:aachen_day_night}
The local features benchmark on the Aachen dataset~\cite{Sattler_2018_CVPR} is restricted to the night part of the dataset. In order to have a more extensive evaluation, we also report the results for the visual localization benchmark on both day and night splits of the dataset. For each query image, the 20 best candidate images are retrieved from the database using NetVlad~\cite{Arandjelovic16} and the official pipeline of the benchmark\footnote{\url{https://github.com/tsattler/visuallocalizationbenchmark}} is then used to estimate the pose of the query images.

\begin{table}[tb]
    \centering
    \caption{\textbf{Visual localization benchmark on the Aachen Day-Night dataset~\cite{Sattler_2018_CVPR}.} We report the percentage of correctly localized queries on both day and night query images for various distance and orientation error thresholds for SIFT, SuperPoint and D2-Net single scale (SS). LISRD leverages illumination variance on the day part and light invariance when querying night images and is thus able to improve the performance of various descriptors on multiple keypoints.}
    \scriptsize
    \setlength{\tabcolsep}{2.6mm}
    \begin{tabular}{cccccccc}
        \toprule
         & \multirow{3}[2]{*}{\makecell{Error\\threshold}} & \multicolumn{2}{c}{SIFT KP} & \multicolumn{2}{c}{SuperPoint KP}  & \multicolumn{2}{c}{D2-Net KP}  \\
        \cmidrule(lr){3-4} \cmidrule(lr){5-6} \cmidrule(lr){7-8}
        & & \makecell{Upright\\Root SIFT} & LISRD & SuperPoint & LISRD & \makecell{D2-Net\\(SS)} & \makecell{LISRD\\(SS)} \\
        \midrule
        \multirow{3}{*}{Day} & $0.5m, 2^\circ$ & 80.8 & \textbf{82.4} & 80.5 & \textbf{85.6} & 77.8 & \textbf{81.7} \\
        & $1m, 5^\circ$ & 89.0 & \textbf{89.9} & 88.3 & \textbf{91.3} & \textbf{89.6} & \textbf{89.6} \\
        & $5m, 10^\circ$ & 93.2 & \textbf{94.2} & 93.2 & \textbf{95.6} & \textbf{95.6} & 94.4 \\
        \midrule
        \multirow{3}{*}{Night} & $0.5m, 2^\circ$ & 51.0 & \textbf{68.4} & 65.3 & \textbf{78.6} & \textbf{78.6} & 71.4 \\
        & $1m, 5^\circ$ & 61.2 & \textbf{79.6} & 73.5 & \textbf{87.8} & \textbf{87.8} & \textbf{87.8} \\
        & $5m, 10^\circ$ & 69.4 & \textbf{91.8} & 86.7 & \textbf{95.9} & \textbf{95.9} & 94.9 \\
        \bottomrule
    \end{tabular}
    \label{tab:aachen_day_night}
\end{table}

Table~\ref{tab:aachen_day_night} shows that LISRD is able to improve over several state-of-the-art descriptors and can generalize to different keypoints. It can indeed leverage variant descriptors for day-day image pairs of the day part and the invariant descriptors on the night part for day-night image pairs.

As the ground truth poses of the Aachen Day-Night dataset have been updated between submission and acceptance of this paper, we additionally show in Table~\ref{tab:aachen_legacy} the results of the local features benchmark on the night part of the Aachen dataset with the old numbers for legacy and easier comparison with previous methods.

\begin{table}[tb]
    \centering
    \caption{\textbf{Legacy results of the local features benchmark on the Aachen Day-Night dataset~\cite{Sattler_2018_CVPR}.} We report the percentage of correctly localized queries for various distance and orientation error thresholds for SIFT, SuperPoint and D2-Net multi-scale (MS). Our method shows a good generalization when evaluated on different keypoints (KP) and can improve the original descriptor performance.}
    \scriptsize
    \setlength{\tabcolsep}{3mm}
    \begin{tabular}{ccccccc}
        \toprule
        \multirow{3}[2]{*}{\makecell{Error\\threshold}} & \multicolumn{2}{c}{SIFT KP} & \multicolumn{2}{c}{SuperPoint KP}  & \multicolumn{2}{c}{D2-Net KP}  \\
        \cmidrule(lr){2-3} \cmidrule(lr){4-5} \cmidrule(lr){6-7}
         & \makecell{Upright\\Root SIFT} & LISRD & SuperPoint & LISRD & \makecell{D2-Net\\(MS)} & \makecell{LISRD\\(MS)} \\
        \midrule
        $0.5m, 2^\circ$ & 33.7 & \textbf{43.9} & 42.9 & \textbf{44.9} & 44.9 & \textbf{45.9} \\
        $1m, 5^\circ$   & 52.0 & \textbf{62.2} & 57.1 & \textbf{65.3} & 64.3 & \textbf{66.3} \\
        $5m, 10^\circ$  & 65.3 & \textbf{82.7} & 77.6 & \textbf{84.7} & \textbf{88.8} & 87.8 \\
        \bottomrule
    \end{tabular}
    \label{tab:aachen_legacy}
\end{table}

\section{Qualitative examples}
\label{sec:qualitative_examples}
\begin{figure}
    \centering
    \newcommand{\sz}{0.48}
    \begin{tabular}{cc}
        \includegraphics[width=\sz\textwidth]{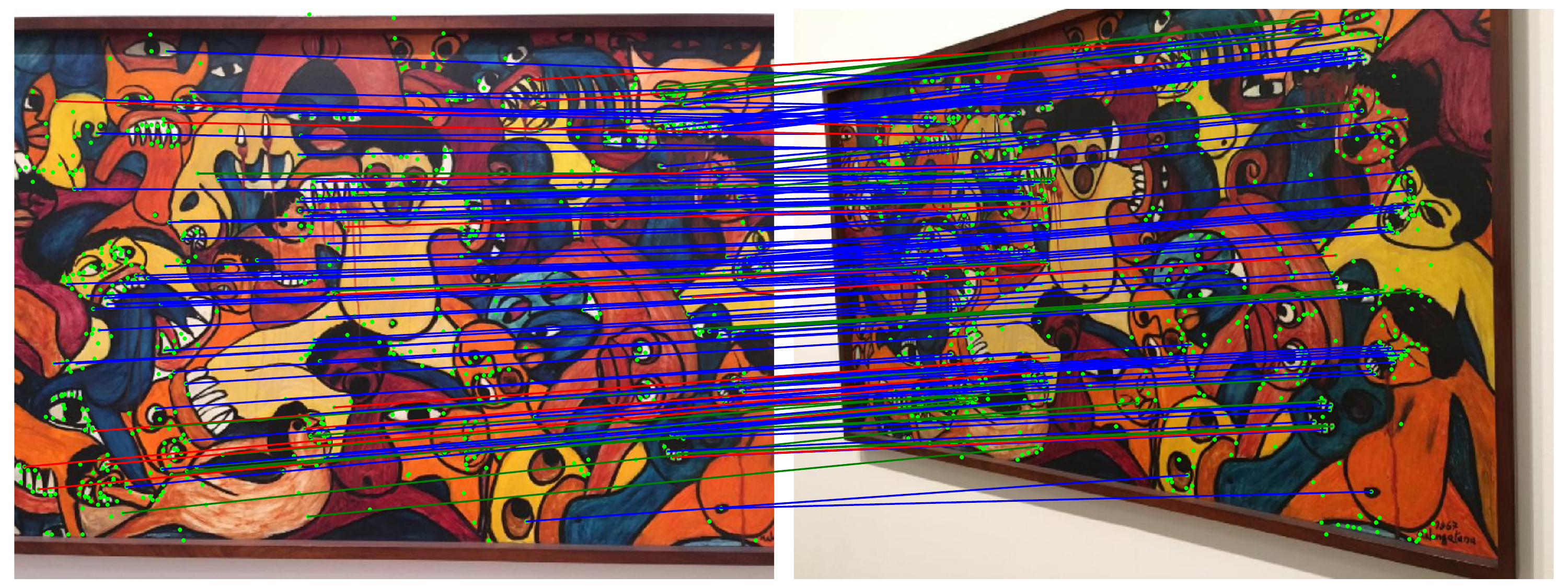}
        & \includegraphics[width=\sz\textwidth]{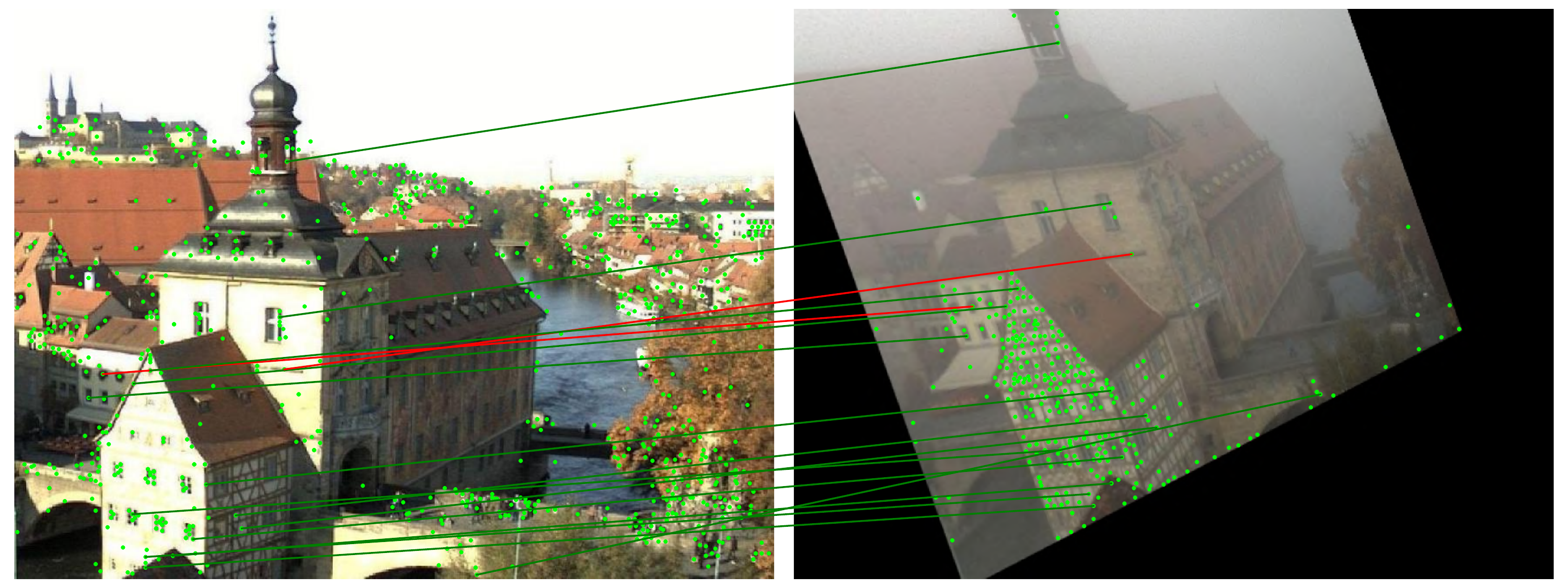} \\
        \includegraphics[width=\sz\textwidth]{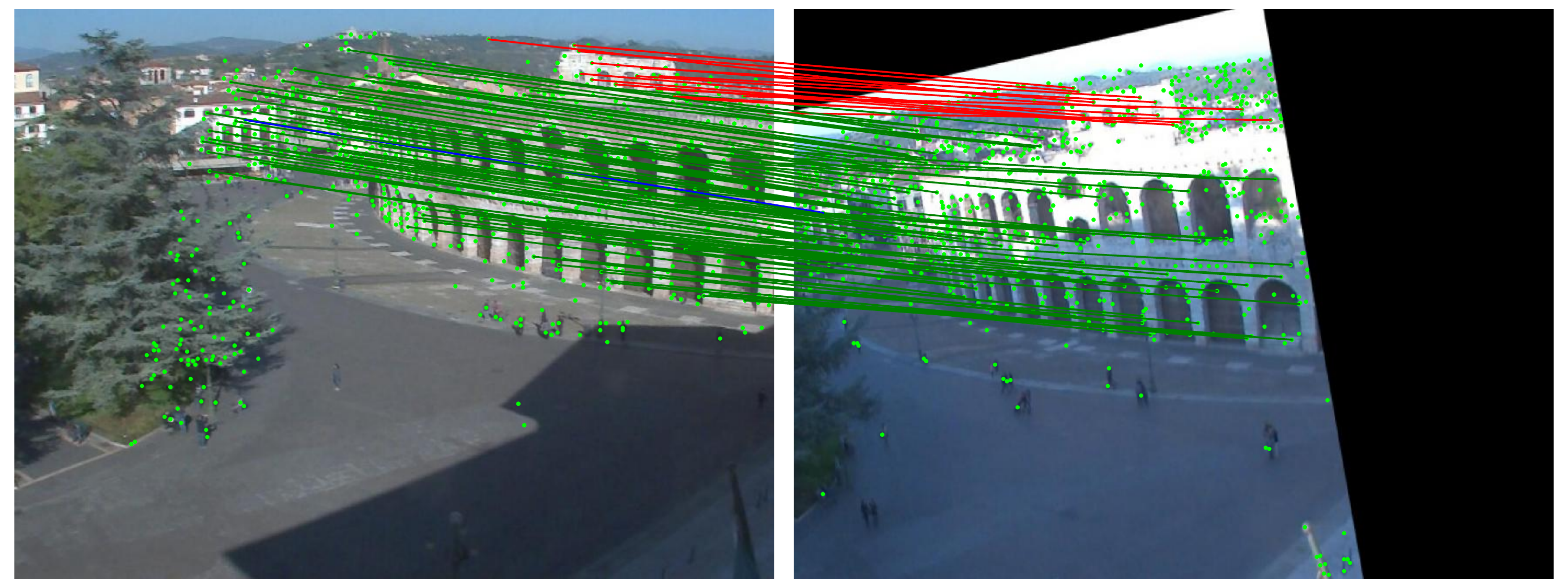}
        & \includegraphics[width=\sz\textwidth]{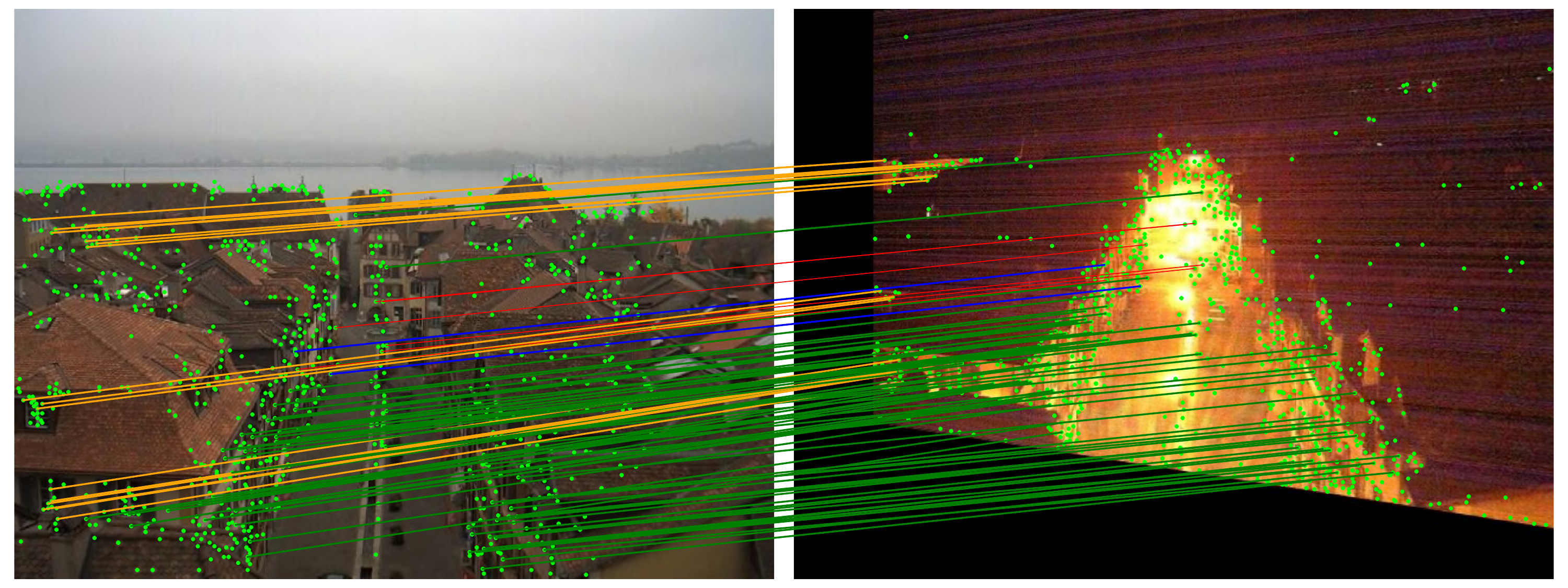} \\
        \includegraphics[width=\sz\textwidth]{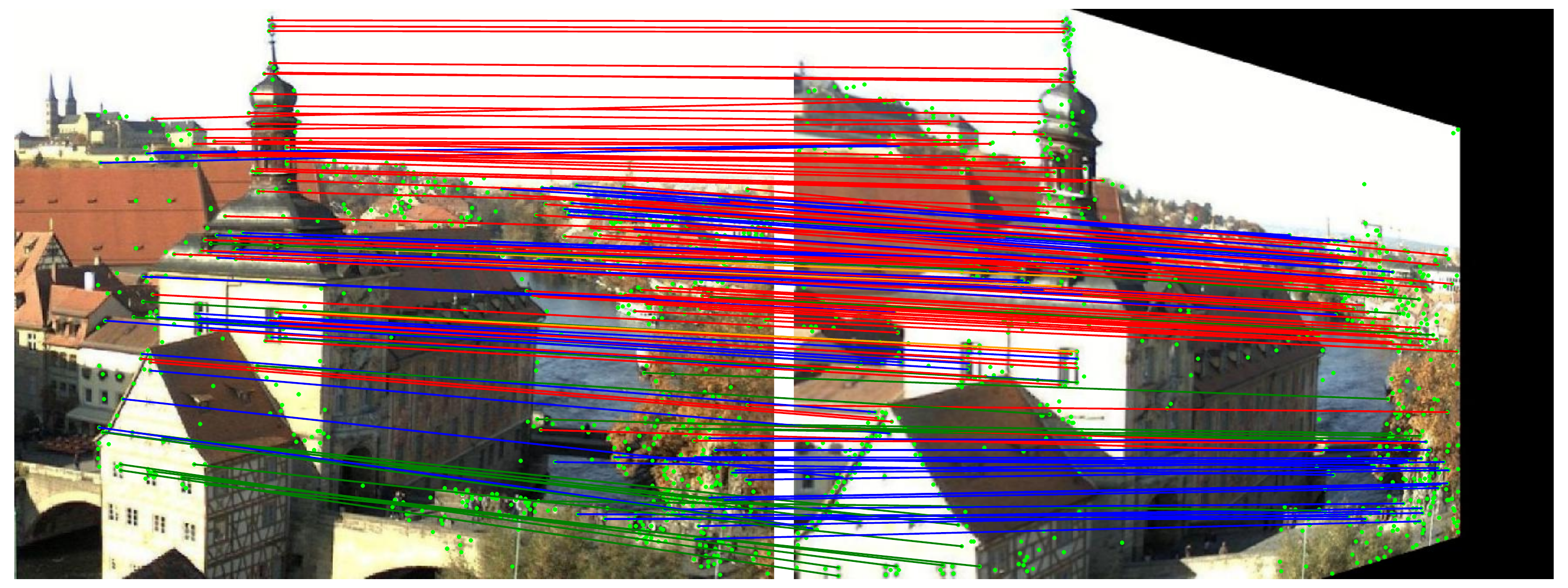}
        & \includegraphics[width=\sz\textwidth]{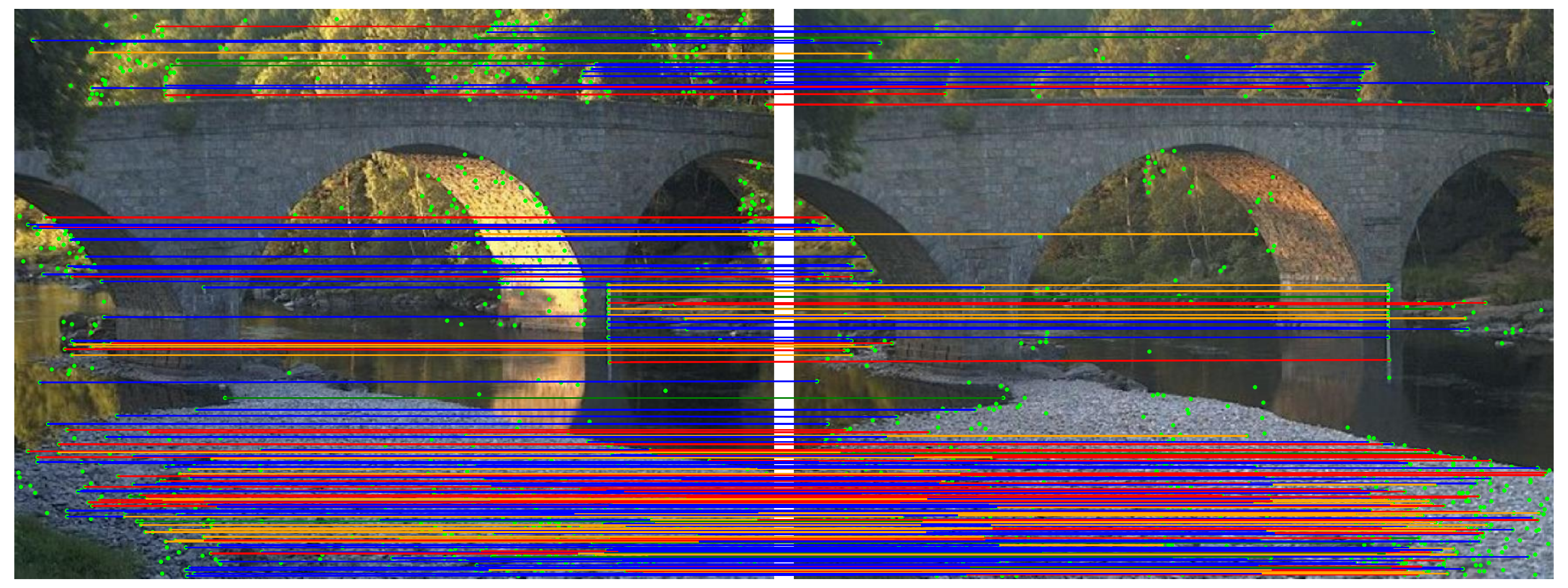} \\
        \multicolumn{2}{c}{\includegraphics[width=0.98\textwidth]{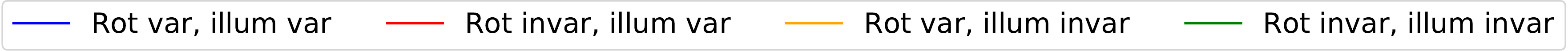}}
    \end{tabular}
    \caption{\textbf{Visualization of the selected invariance.} Matches of SIFT keypoints with LISRD descriptors are filtered with mutual nearest neighbor and RANSAC~\cite{ransac}. Since our method uses a soft weighting of the invariances, each color corresponds only to the invariance that contributed the most to validate the match. First line: one type of invariance predominates in the whole image. Second line: two invariances are relevant in the same image (on the left, rotation invariance is needed, but the building in the top right corner is overexposed in both images and illumination invariance is not needed in this area ; on the right, illumination invariance is needed, but the image is upright on the left side, while the distortion creates a rotation on the central part and rotation invariance becomes necessary). Third line: multiple different invariances can be leveraged in the same image (on the left image, the right part of the image is mainly upright and with constant illumination, while the house in the lower left corner is overexposed and rotated, hence the fully invariant descriptor is selected ; on the right image, most of the selected descriptors are rotation variant as the viewpoint is fixed, but the left pier of the bridge has a constant illumination while the right pier has a different illumination and the illumination invariant descriptor predominates).}
    \label{fig:4descs_vis}
\end{figure}

We provide additional qualitative examples of matches based on SIFT keypoints and LISRD descriptors. All matches are filtered with mutual nearest neighbor, followed by a homography fitting with RANSAC~\cite{ransac}. Figure~\ref{fig:4descs_vis} brings a visualization of the invariance selection, with a different color for each kind of invariance that was selected. Since the selection is in practice based on a soft weighting, we only show the color of the learned descriptors that contributed the most in the matching decision. These sample images show that in some situations, a single invariance is sufficient for the full image, but in other cases multiple invariances can be leveraged within the same image, demonstrating the need of tiled meta descriptors. This is for example useful when the overall illumination is constant in a pair of images, but one part an image (e.g. a building) is overexposed or in the shadow.

\begin{figure}
    \centering
    \newcommand{\sz}{0.48}
    \begin{tabular}{cc}
        \includegraphics[width=\sz\textwidth]{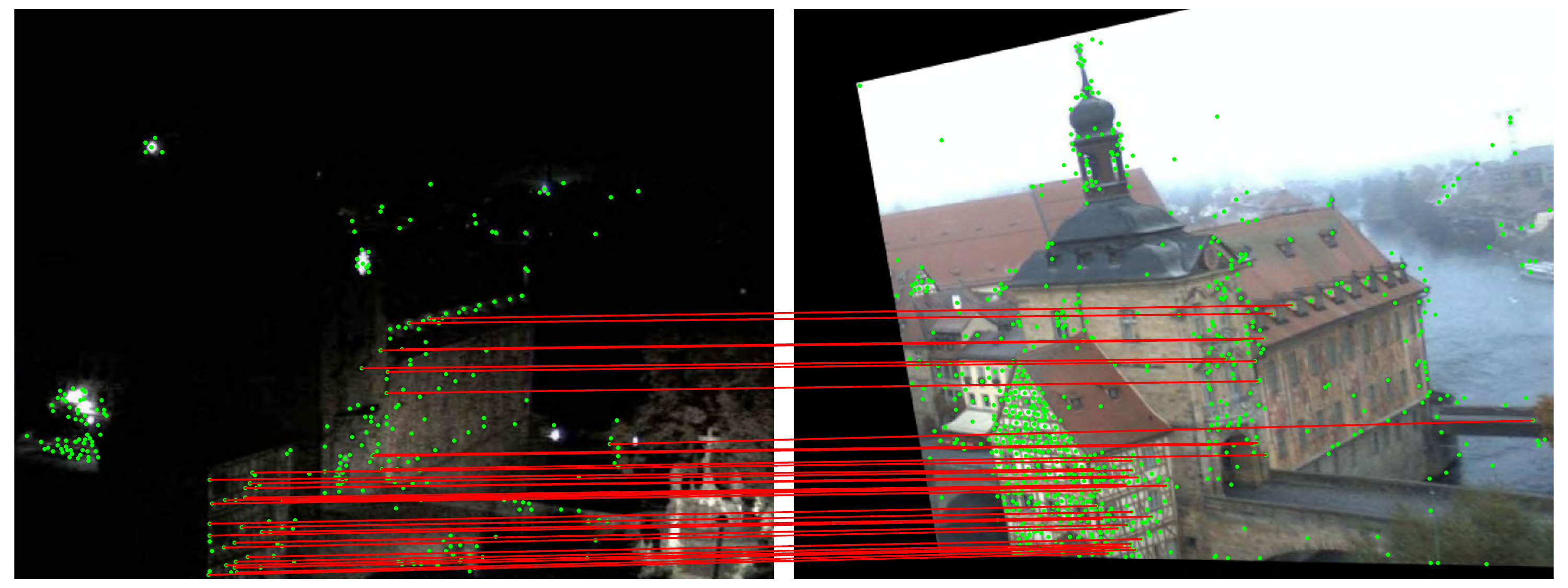}
        & \includegraphics[width=\sz\textwidth]{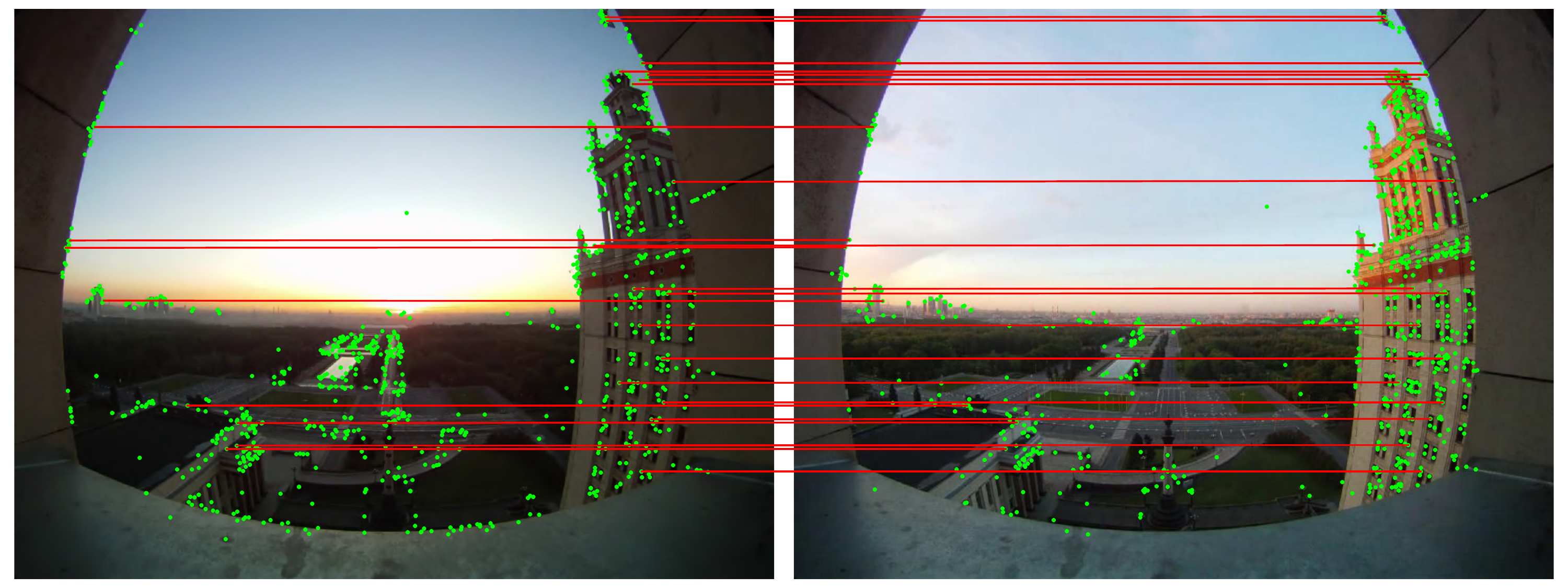} \\
        \includegraphics[width=\sz\textwidth]{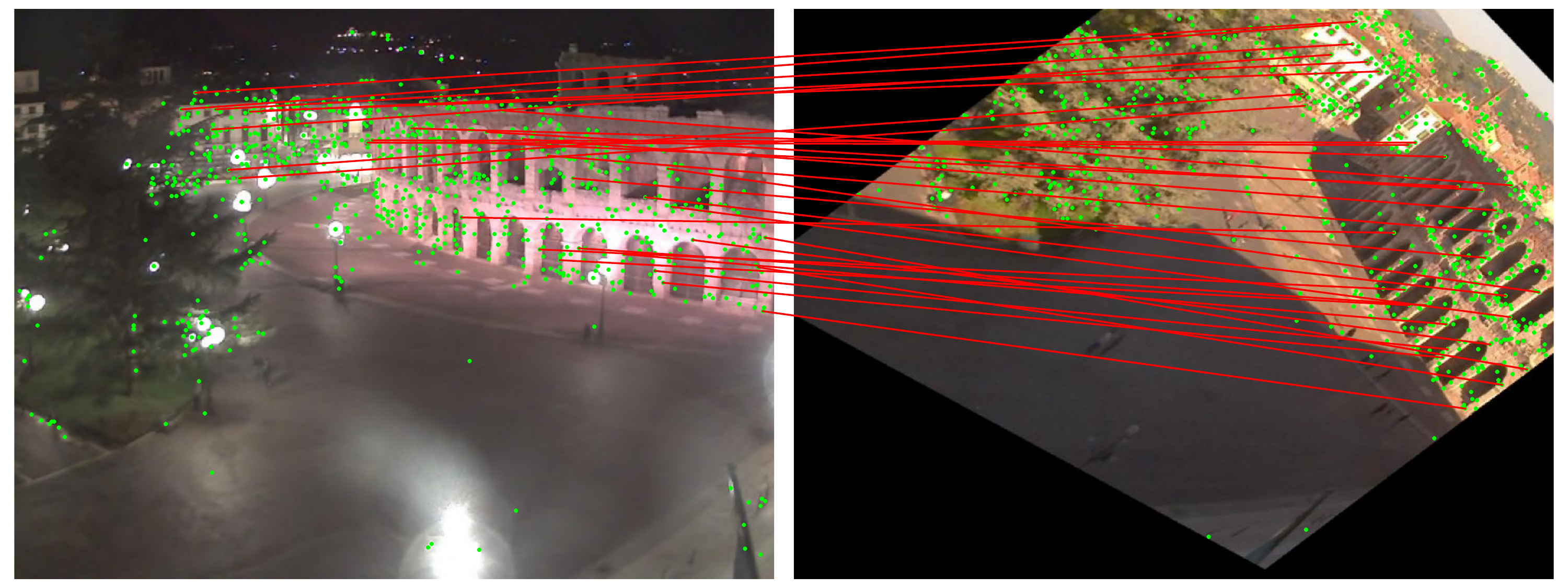}
        & \includegraphics[width=\sz\textwidth]{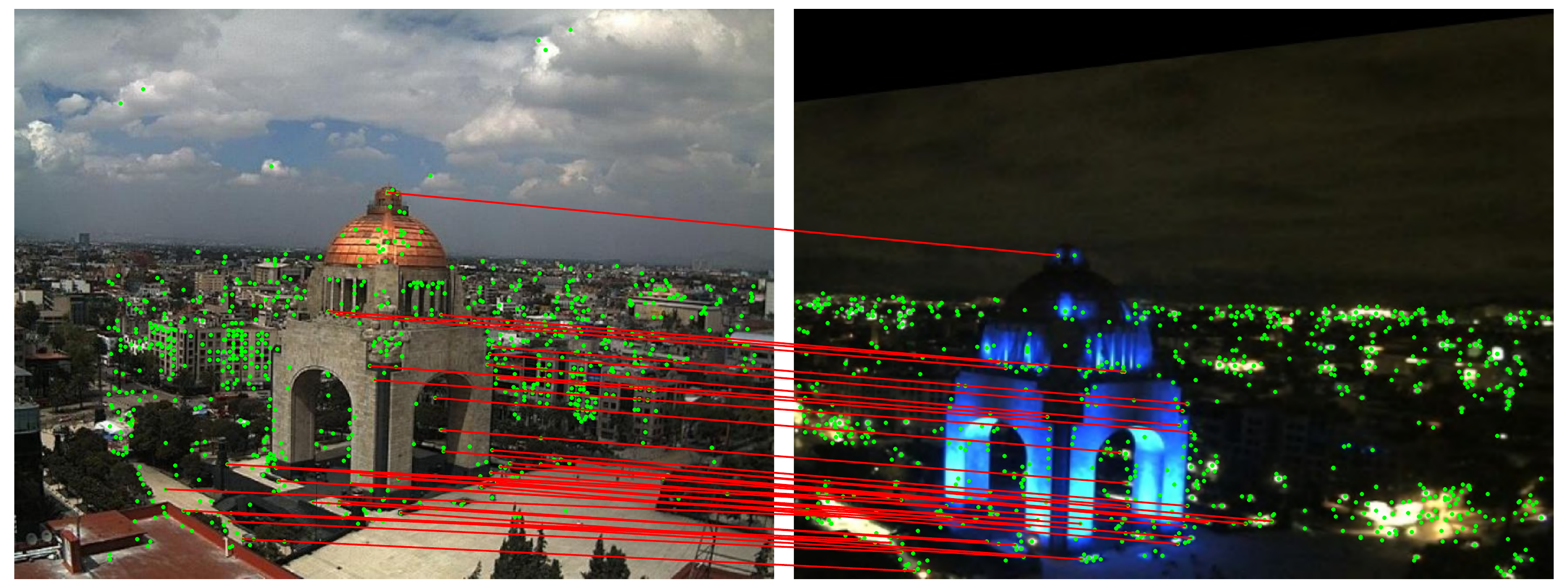} \\
        \includegraphics[width=\sz\textwidth]{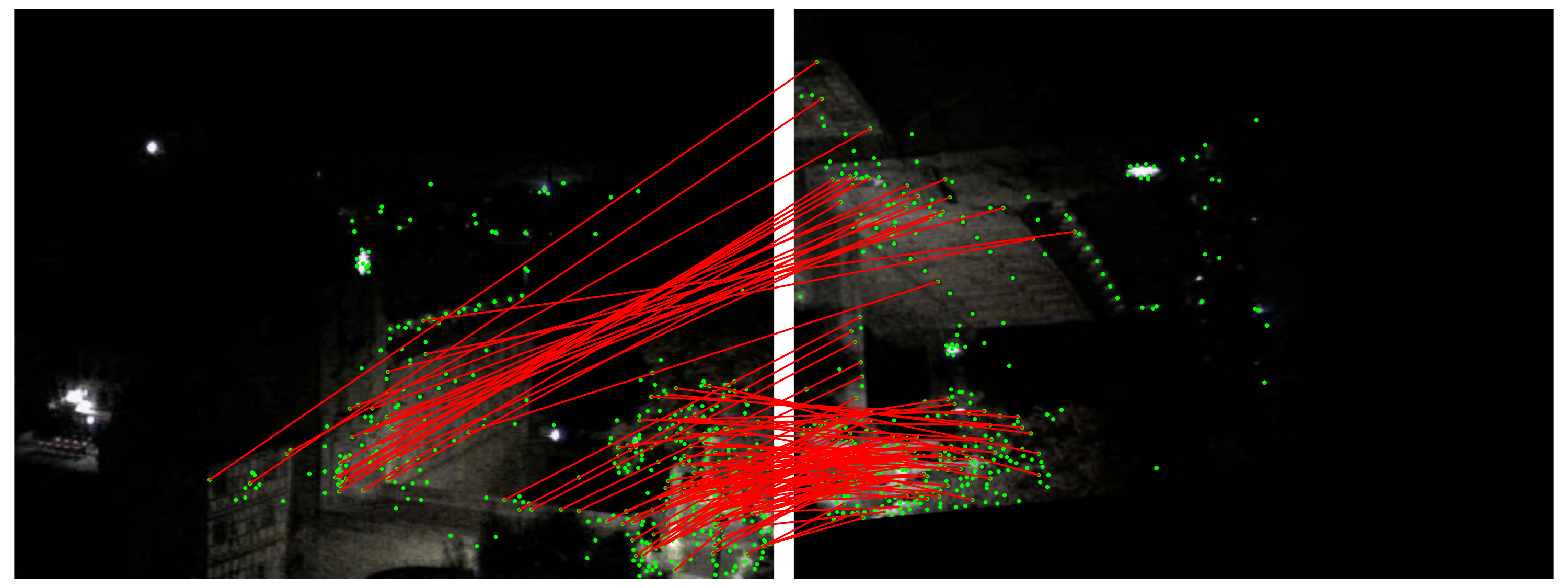}
        & \includegraphics[width=\sz\textwidth]{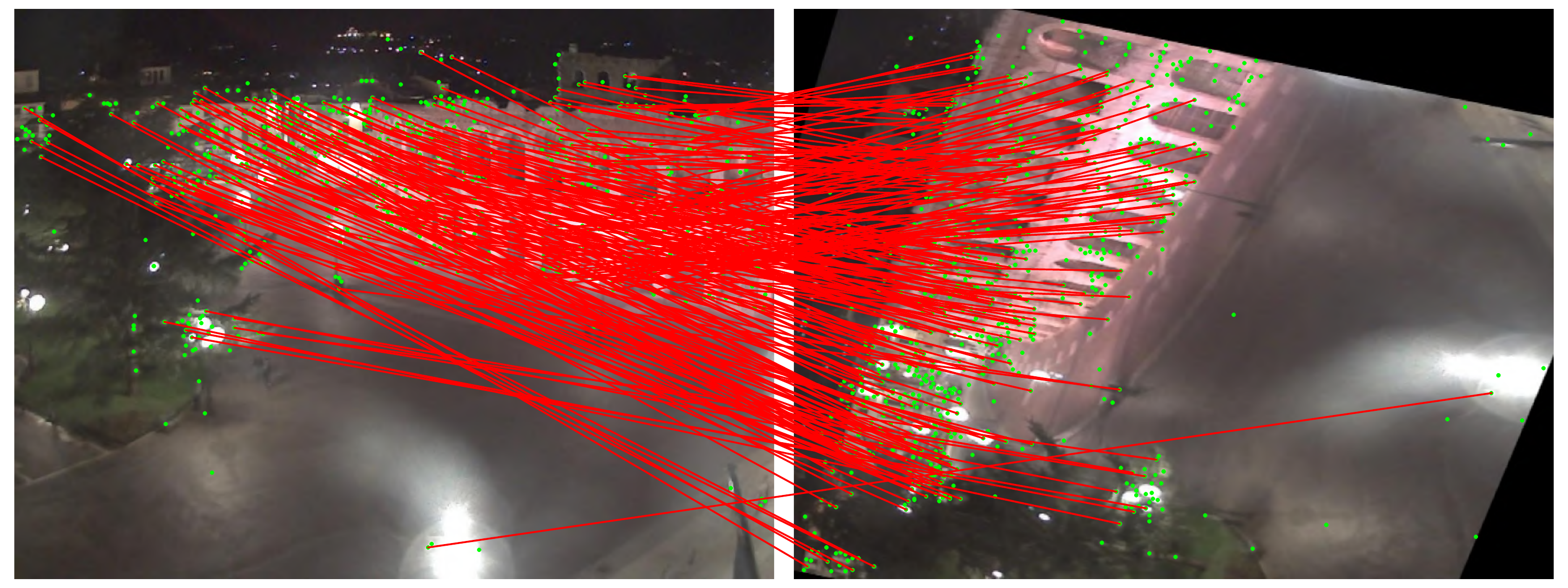}
    \end{tabular}
    \caption{\textbf{Matches in challenging situations.} SIFT keypoints are detected, matched with the LISRD descriptors, and mutual nearest neighbor and RANSAC~\cite{ransac} are used to filter out wrong matches. A single red color is used for all the inlier matches, regardless of the chosen invariance. Matches based on LISRD descriptors are able to handle strong illumination changes such as day-night, inter-image illumination variations in day-day and night-night pairs, and small as well as strong rotations.}
    \label{fig:challenging_scenes}
\end{figure}

Finally, Figure~\ref{fig:challenging_scenes} displays a selection of matches in challenging scenarios, for example with day-night and/or with strongly rotated images.

\clearpage
\bibliographystyle{splncs04}
\bibliography{supplementary}